\pdfoutput=1  % arXiv: build with pdflatex (all figures are PDF)
\documentclass[sigconf,nonacm,10pt]{acmart}
\usepackage{amsthm}
\usepackage{listings}
\usepackage{lstlinebgrd}  % linebackgroundcolor, used by the appendix listings
\usepackage{cleveref}
\usepackage{multirow}
\usepackage{xspace}
\usepackage{soul}
\usepackage{tikz}

\graphicspath{{figures/}}

\def\conference
\usepackage{etoolbox}
\newcommand{\ifconference}[1]{{{\ifx\fullversion\undefined{#1}\fi}\xspace}}
\newcommand{\iffullversion}[1]{{{\ifx\conference\undefined{#1}\fi}\xspace}}

\usepackage{graphicx}  % pictures and figures
\usepackage{lipsum}  % random paragraphs
\newcommand{\hide}[1]{} % hide
\usepackage{xspace}
\usepackage{textcomp}
\usepackage{comment} % \begin{comment} ... \end{comment}
\usepackage{verbatim}
\definecolor{mypink1}{rgb}{0.858, 0.188, 0.478}
\definecolor{mypink2}{RGB}{219, 48, 122}
\definecolor{mypink3}{cmyk}{0, 0.7808, 0.4429, 0.1412}
\definecolor{mygray}{gray}{0.6}
\usepackage{url}

\newcommand{\emp}[1]{\emph{\textbf{#1}}} % highlight
\let \originalleft \left
\let\originalright\right
\renewcommand{\left}{\mathopen{}\mathclose\bgroup\originalleft}
\renewcommand{\right}{\aftergroup\egroup\originalright}

\usepackage{scalerel} % stretch a formula

\newtheoremstyle{exampstyle}
{.5em} % Space above
{1em} % Space below
{\it} % Body font
{.5em} % Indent amount
{\it \bfseries} % Theorem head font
{.} % Punctuation after theorem head
{.5em} % Space after theorem head
{} % Theorem head spec (can be left empty, meaning `normal')
\theoremstyle{exampstyle} 
\theoremstyle{exampstyle} 
\theoremstyle{exampstyle} 
\theoremstyle{exampstyle} 

\makeatletter

\newcommand{\whp}[1]{\emph{whp}}

\usepackage{pifont}
\usepackage[shortlabels]{enumitem}

\setlist{topsep=0.3em,itemsep=0.2em,parsep=0.1em,leftmargin=*}
\usepackage{float}
\usepackage[labelfont=bf,font={small},aboveskip=0em, belowskip=0em]{caption}

\usepackage[labelfont=bf,list=true,skip=0em]{subcaption}
\usepackage[rightcaption]{sidecap}

\usepackage{wrapfig}

\usepackage{array}% for extended column definitions
\newcolumntype{L}[1]{>{\raggedright\let\newline\\\arraybackslash\hspace{0pt}}m{#1}}
\newcolumntype{C}[1]{>{\centering\let\newline\\\arraybackslash\hspace{0pt}}m{#1}}
\newcolumntype{R}[1]{>{\raggedleft\let\newline\\\arraybackslash\hspace{0pt}}m{#1}}
\newcolumntype{B}{>{\bf}c}
\usepackage{rotating}

\usepackage{booktabs} % provides toprule, bottomrule, midrule, cmidrule, etc.
\usepackage{multicol,multirow}
\usepackage{longtable} % provides long table
\usepackage{supertabular} % similar to long table, allowing tables to take more than one page
\usepackage{colortbl}
\usepackage{bigstrut}
\usepackage{titlesec}
\newcommand{\mysubsubsection}[1]{{#1}.}
\titleformat{\subsubsection}[runin]
{\normalfont\normalsize\bfseries}{\thesubsubsection}{1em}{\mysubsubsection}

\newcommand{\myparagraph}[1]{\vspace{0.2em}\noindent\emp{#1}~~}

\usepackage[algo2e,ruled,lined,linesnumbered,noend]{algorithm2e}
\usepackage[noend]{algpseudocode}

\makeatletter
\patchcmd{\@algocf@start}% <cmd>
  {-1.5em}% <search>
  {0pt}% <replace>
  {}{}% <success><failure>
\newcommand{\nosemic}{\renewcommand{\@endalgocfline}{\relax}}% Drop semi-colon ;
\newcommand{\dosemic}{\renewcommand{\@endalgocfline}{\algocf@endline}}% Reinstate semi-colon ;
\SetSideCommentLeft
\SetKwInput{notations}{Notations}
\SetKwInput{notes}{Notes}
\SetKwInput{maintains}{Maintains}

\SetKwProg{myfunc}{Function}{}{}
\SetKwFor{parForEach}{ParallelForEach}{do}{endfor}
\SetKwFor{Justrepeat}{Repeat}{}{}

\SetKw{MIN}{min}
\SetKw{MAX}{max}
\SetKw{OR}{or}
\SetKw{AND}{and}

\SetCommentSty{mycommfont}

\usepackage{mdframed}
\definecolor{framelinecolor}{RGB}{68,114,196}
\mdfdefinestyle{mystyle}{linecolor=framelinecolor,innertopmargin=1pt,innerbottommargin=2pt,backgroundcolor=gray!20,skipabove=2pt,skipbelow=0pt}%leftmargin=0,topmargin=
\mdfdefinestyle{densestyle}{linecolor=framelinecolor,innertopmargin=0,innerbottommargin=0,leftmargin=0,rightmargin=0,backgroundcolor=gray!20}
\mdfdefinestyle{compactcode}{linecolor=framelinecolor,innertopmargin=1pt,innerbottommargin=1pt,backgroundcolor=gray!20,skipabove=0pt,skipbelow=0pt,leftmargin=0,rightmargin=0}

\usepackage{framed}

\usepackage{listings}

\newdimen\zzsize
\newdimen\kwsize
\newcommand{\basicstyle}{\fontsize{\zzsize}{1\zzsize}\ttfamily}
\newcommand{\keywordstyle}{\fontsize{\kwsize}{1\kwsize}\ttfamily\bf}

\newdimen\zzlstwidth
\usepackage{cleveref}
\crefname{section}{Sec.}{Sec.}
\crefname{theorem}{Thm.}{Thm.}
\crefname{lemma}{Lem.}{Lem.}
\crefname{corollary}{Col.}{Col.}
\crefname{table}{Tab.}{Tab.}
\crefname{algorithm}{Alg.}{Alg.}
\crefname{figure}{Fig.}{Fig.}
\crefname{fact}{Fact}{Fact}
\Crefname{table}{Tab.}{Tab.}
\crefname{problem}{Problem}{Problem}

\usepackage{tikz} % draw geometric objects

\definecolor{dpcol}{RGB}{0,160,240}

\newcommand{\yan}[1]{{\color{violet}{\bf Yan:} #1}}
\newcommand{\yihan}[1]{{\color{blue}{\bf Yihan:} #1}}

\newcommand{\zheqi}[1]{{\color{dpcol}{\bf Zheqi:} #1}}

\newcommand{\complete}[1]{}

\newcommand{\Nu}{\mathcal{N}}

\newcommand{\knn}{$k$-NN}

\newcommand{\pred}{\mathcal{P}}

\crefname{section}{Section}{Section}
\crefname{@theorem}{Theorem}{Theorem}
\crefname{table}{Table}{Table}
\crefname{figure}{Figure}{Figure}
\crefname{algocf}{Alg.}{Alg.}
\Crefname{algocf}{Algorithm}{Algorithms}

\definecolor{dpcoldark}{RGB}{0,80,120}
\definecolor{codeorange}{RGB}{120,64,16}
\definecolor{codegreen}{rgb}{0,0.5,0}
\definecolor{codegray}{rgb}{0.5,0.5,0.5}
\definecolor{codepurple}{rgb}{0.58,0,0.82}
\definecolor{backcolour}{rgb}{0.95,0.95,0.92}
\definecolor{commcol}{RGB}{0,127,0}

\lstdefinestyle{mystyle}{
  commentstyle=\color{codegreen},
  keywordstyle=\color{dpcoldark},
  numberstyle=\tiny\color{codegray},
  stringstyle=\color{codeorange},
  basicstyle=\ttfamily\small\linespread{0.8}\selectfont,
  emphstyle=\underbar,
  breakatwhitespace=false,
  breaklines=true,
  captionpos=b,
  keepspaces=true,
  numbers=left,
  numbersep=5pt,
  showspaces=false,
  showstringspaces=false,
  showtabs=false,
  tabsize=2,
  frame=single
}
\newcommand{\option}[1]{{\color{lightgray}\textit{(#1)}}}
\newcommand{\annlib}{ANNLib}
\newcommand{\Bsearch}{\texttt{Collect}}
\newcommand{\bsearch}{\texttt{collect}}
\newcommand{\prune}{\texttt{prune}}
\newcommand{\robustprune}{\textit{robustPrune}}
\newcommand{\fnbhs}[1]{\texttt{f\_nbhs#1}}
\newcommand{\fdist}[1]{\texttt{f\_dist#1}}
\renewcommand{\pred}[1]{\texttt{pred#1}}
\newcommand{\agent}{\textit{edge agent}}

\newcommand{\pforeach}[1]{\textit{parallel-for-each#1}}
\newcommand{\vamana}{\textit{Vamana}}
\newcommand{\hnsw}{\textit{HNSW}}
\newcommand{\annlibvamana}{ANNLib-Vamana}
\newcommand{\annlibhnsw}{ANNLib-HNSW}
\newcommand{\parlayvamana}{Parlay-Vamana}
\newcommand{\parlayhnsw}{Parlay-HNSW}
\newcommand{\svamana}{\textit{Stitched-Vamana}}
\newcommand{\fvamana}{\textit{Filtered-Vamana}}
\newcommand{\fprune}{\textit{filtered prune}}
\newcommand{\parray}{\textit{prefix-shared array}}

\makeatletter
\def\@authorfont{\Large}
\def\@affiliationfont{\normalsize}
\makeatother

\begin{document}
%-------------------------------------------------------------------------------

\title{{\huge ANNLib: A Development Framework for Efficient Approximate Nearest Neighbor Search}}

\settopmatter{authorsperrow=5}

\author{Zheqi Shen}
\affiliation{\institution{UC Riverside}\country{Riverside CA, USA}}
\email{zshen055@ucr.edu}
\author{Zijin Wan}
\affiliation{\institution{UC Riverside}\country{Riverside CA, USA}}
\email{zwan019@ucr.edu}
\author{Jingbo Su}
\affiliation{\institution{W\&M}\country{Williamsburg VA, USA}}
\email{jsu02@wm.edu}
\author{Yan Gu}
\affiliation{\institution{UC Riverside}\country{Riverside CA, USA}}
\email{ygu@cs.ucr.edu}
\author{Yihan Sun}
\affiliation{\institution{UC Riverside}\country{Riverside CA, USA}}
\email{yihans@cs.ucr.edu}

\renewcommand{\shortauthors}{}

%% Short names used by the VLDB reference block.
\renewcommand{\shorttitle}{ANNLib: A Development Framework for Efficient ANNS}

%-------------------------------------------------------------------------------
\begin{abstract}
%-------------------------------------------------------------------------------
Approximate Nearest Neighbor Search (ANNS) plays a pivotal role in modern deep learning pipelines. Recently, many ANNS systems have been proposed to provide broad, flexible functionalities or achieve high performance. However, it is inherently difficult to achieve both.

We propose \textbf{\annlib{}} to address this gap.
\annlib{} is a library that provides a programming framework to achieve high performance and flexible functionalities for ANNS systems, based on popular graph-based ANNS algorithms.
We carefully decouple and independently optimize both the \emph{algorithm} and the \emph{data structure} components in an ANNS system.
In addition, we integrate state-of-the-art algorithms and data structures as modules in \annlib{}, as well as our new designs.
Users can choose combinations of components to support sophisticated settings with high performance, such as filtered search, fully dynamic updates, historical queries on snapshots, and range searches. Our experiments show that our new solution provides a simple interface for various applications, and achieves comparable or even better performance to previous work specifically for each application.
\end{abstract}

\maketitle

%-------------------------------------------------------------------------------
% Main content sections
%-------------------------------------------------------------------------------
\section{Introduction}

Approximate nearest neighbor search (ANNS) is essential for efficient data retrieval in modern deep learning pipelines.
%After the training phase, objects are embedded into high-dimensional vector spaces, where many application-specific queries reduce to a nearest neighbor search within this space. 
Typically, models embed objects into high-dimensional spaces, casting diverse downstream queries as \emph{$k$-nearest neighbor searches (\knn{} search)}. 
%Formally, the problem is known as \emph{$k$-nearest neighbor (\knn{}) search}, which takes a set of high-dimensional points $P\subseteq \mathbb{R}^d$ and a query point $q\in\mathbb{R}^d$, and returns the $k$ closest points to $q$ in $P$ based on some distance measurements. 
Formally, given a dataset $P\subseteq \mathbb{R}^d$ and a query point $q\in\mathbb{R}^d$, a \knn{} search returns the $k$ closest points to $q$ in $P$ under a given distance metric. 

Given the inherent difficulty of searching in high dimensions, approximate methods are used for practical efficiency. 
Over the past decade, ANNS has emerged as a prominent research focus. It has driven the development of many algorithms~\cite{douze2024faiss,subramanya2019diskann,malkov2020hnsw,munoz2019hcnng,ren2020hmann,chen2021spann,pham2022falconn}, most notably graph-based approaches, and enabled a wide range of applications~\cite{chase2023vector,tawalke2023pr,bing,stallbaumer2023copilot,eirinaki2018recommender,lecun1995convolutional,mikolov2013distributed,scarselli2008graph,tagami2017annexml,zhang2022uni,zhang2018visual,huang2020embedding}.
% Over the past decade, ANNS has emerged as a prominent research focus, driving the development of many algorithms~\cite{douze2024faiss,subramanya2019diskann,malkov2020hnsw,munoz2019hcnng,ren2020hmann,chen2021spann,pham2022falconn} --- most notably graph-based approaches---and enabling a wide range of applications~\cite{chase2023vector,tawalke2023pr,bing,stallbaumer2023copilot,eirinaki2018recommender,lecun1995convolutional,mikolov2013distributed,scarselli2008graph,tagami2017annexml,zhang2022uni,zhang2018visual,huang2020embedding}.
%A graph-based ANNS algorithm builds a proximity graph, where each point in the dataset is a vertex. 
%Every point is connected to other geometrically-close points, as well as some ``long edges'' to keep the graph connected. 
Graph-based methods index the dataset as a proximity graph, where vertices (data points) are connected to their spatial neighbors, alongside some ``long edges'' to maintain global connectivity.
An ANNS query traverses the graph from an entry point using a \emph{beam search}, greedily moving toward the query vector to identify the $k$ nearest neighbors. 
Graph-based algorithms have become the dominant choice for ANNS systems due to their ability to achieve high recall (the fraction of true \knn{s} identified) while maintaining efficient throughput (queries per second, or QPS). 
% \textbf{In this paper, we study graph-based ANNS system that support both \emph{a broad range of functionalities} and \emph{strong performance}.}
\textbf{In this paper, we study graph-based ANNS systems that combine \emph{a broad range of functionalities} with \emph{strong performance}.}
Given its broad applicability, ANNS is widely deployed~\cite{yu2025pecann,dhulipala2024muvera}. Beyond the standard static setting, different scenarios demand specialized \emph{functionalities}.
For example, search and recommendation engines are required to handle frequent \emph{updates} to the index~\cite{xu2023spfresh,li2018JD,zhang2025cleann};
shopping platforms require \emph{filters} over product attributes for item search~\cite{gollapudi2023filtered,guo2022manu,wang2021milvus};
analytical workloads may need historical \emph{snapshots} for temporal queries~\cite{wei2020analyticdb}.
%Depending on the application, many ANNS systems require these (or more) functionalities.
As a result, a developer must consider \emph{both the choice of algorithm and whether (and how) it can be adapted to different functionalities in the target application.}
%its compatibility with the required functionalities.

%Beyond functionality, as data volumes grow---both in size and dimensions---\emp{performance} becomes increasingly critical for ANNS systems.
%For example, the real-world datasets used in this paper in \cref{exp:dataset}, where the OpenAI~\cite{neelakantan2022text} dataset contains 5 million points in 1,536 dimensions, and BIGANN contains up to billions of points. 
%Such large-scale data necessitates high performance in ANNS systems. 
Beyond functionalities, the ever-growing scale and dimensionality of modern datasets (e.g., the 1{,}536-dimensional OpenAI~\cite{neelakantan2022text} and the billion-scale BIGANN~\cite{amsaleg2010datasets} in \cref{exp:dataset}) make \emph{performance} increasingly critical for ANNS. 
This demand applies not only to queries but also to construction, updates, and other functionalities mentioned above, all of which often require \emph{parallelism}. 
Consequently, a developer must consider \emph{both high-level algorithmic choices and low-level, performance-critical implementation details.} 
As ANNS continues to attract attention,
research efforts have largely focused on either broadening functionality or boosting performance.
Despite significant progress on both fronts, few systems excel at both.
Feature-rich vector databases~\cite{douze2024faiss,pinecone,weaviate,lucene,wei2020analyticdb} prioritize an easy-to-use interface over efficiency,
but underperform specialized algorithms.
Conversely, performance-optimized systems~\cite{dobson2022parallel,hnswlib2019hnswlib,subramanya2019diskann,munoz2019hcnng,singh2021fresh,men2025pkdtree,ding2026skiplist} consist of code refined over years with sophisticated optimizations,
making them hard to adapt to new functionalities, even for experts.
%Such databases maintain an index for the input points, and users can easily realize different queries, 
%such as filter queries and updates. 
%However, Their main target are end users rather than developers---they provide easy interface for users to describe the needs, but not the capability to modify the algorithms and data structures implementing them.

In this paper, we propose \textbf{\annlib{}} to address this gap.  
\annlib{} is a library providing a programming framework to achieve high performance and flexible functionalities for ANNS systems, based on popular graph-based ANNS algorithms. 
The design goal of \annlib{} is to \emp{ease the development} of an ANN system while \emp{maintaining high performance}. 
Importantly, \annlib{} encapsulates core algorithmic primitives for ANNS, all natively supporting \emp{high parallelism}. 
By design, \annlib{} decouples, modularizes, and independently optimizes the \emph{algorithm} and \emph{data structure} components of an ANNS system, 
%allowing for different combinations and extensions based on the requirement of applications. 
%In particular, \annlib{} strategically separates \emph{algorithms} from \emph{data structures} within the ANNS system, 
% allowing for flexible combinations and extensions based on the requirements of applications. 
enabling flexible combinations and extensions based on application requirements. 
% Each component offers several deeply-optimized, pre-built implementations, while also supporting custom implementations from users through a standard interface. 
Each component offers several optimized building blocks and provides multi-level interfaces for adequate customization. 
%This enables flexible functionality and overall efficiency for different use scenarios. 
For algorithms, \annlib{} provides basic primitives such as parallel beam search and batch insertions, which are useful in various graph-based ANNS. 
It also provides pre-built graph-based algorithms such as \vamana{}~\cite{subramanya2019diskann} and HNSW~\cite{malkov2020hnsw}. 
For data structures, \annlib{} abstracts a concise interface for algorithms to interact with the underlying graph. 
We implemented several pre-built data structures for the graph index in different use scenarios, such as Compressed Sparse Row (CSR) for static settings, 
and functional trees~\cite{dhulipala2022pac} for streaming and/or historical snapshot settings. 
We also proposed a new data structure optimized for the features of ANNS graphs with dynamic updates.

With the primitives and data structures in \annlib{}, complex ANNS functionalities can be easily constructed through modular combinations. Notably, it allows general algorithmic ideas to be implemented with minimal effort while enabling natural parallelism. 
To demonstrate \annlib{}'s flexibility and effectiveness, we build solutions for several applications: (1) \emph{fully dynamic updates} with batch insertions and deletions \cite{singh2021fresh}; (2) \emph{filtered search} by implementing \svamana{} and \fvamana{} \cite{gollapudi2023filtered}; (3) \emph{streaming updates} with historical snapshots, and (4) \emph{range-filter} search for attribute range search. 
Taking the range-filter search as a case study, we illustrate how \annlib{} simplifies the implementation of existing algorithms and improves them via our pluggable components. 
Our first example is based on \emph{UNIFY}, originally bottlenecked by long construction time; \annlib{} natively enables parallelism for UNIFY, outperforming its original version with naive parallelism. 
The second example is \emph{WinFilter}, a parallel solution with bottleneck of space usage. 
Re-implementing it with \annlib{} enables flexible integration of alternative parallel data structures in \annlib{}, substantially reducing space consumption. 

\hide{
We carefully evaluate the implementations based on \annlib{}. 
We compare all implementations in \annlib{} with existing implementations specifically optimized for these applications.
}
We evaluate \annlib{}'s implementations against existing work specifically optimized for these applications.
In most cases, \annlib{} achieves more competitive or superior performance (construction time and query throughput) and quality (query recall) compared to existing solutions, while allowing for a much more concise implementation. 
We note that based on our design, \annlib{} can also support more combinations of applications. 
For example, using the algorithms for filtered search and the data structure for historical snapshots, one can also perform filtered queries on a snapshot of a dynamically changed dataset. 
% We release our code at ~\cite{annlib-pub}, and put more details and experiments in the supplemental material~\cite{??}.
We release our code at ~\cite{annlib-pub}.

\begin{figure}[t]
  \centering
  \includegraphics[width=\linewidth]{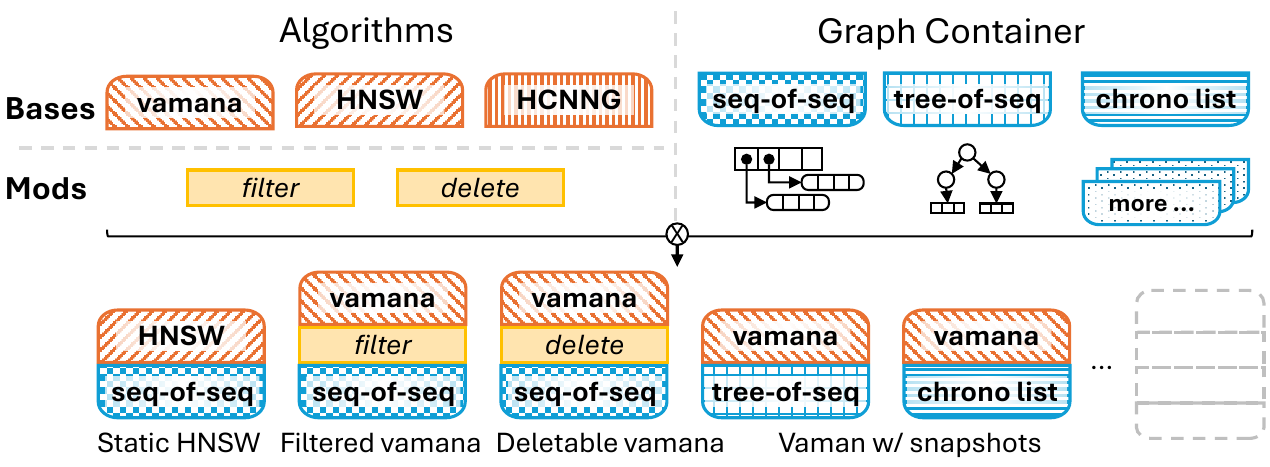}
  \caption{\textbf{An overview of \annlib{}.} 
  Users can combine different components, including the base algorithm, algorithm modules and the graph container data structure, as needed by their applications.  }
  \label{fig:mod-demo}
\end{figure}

\section{Background on ANNS}\label{sec:background}
% describe the common design most graph-based ANN algorithms follow, break down their algorithms and analyze the common parts. Throughout the analysis, we identify the hinderer that prevent the existing systems from being efficiently modularized.
%In this section, we briefly describe the background about ANNS. More related work is reviewed in \cref{sec:related-work}.

\subsection{Problem Definitions}
%\yan{use myparagraph}
%\myparagraph{ANN Search.}
Given a set $P\subseteq\mathbb{R}^d$ and a query $q\in \mathbb{R}^d$, the $k$-nearest neighbor search ($k$-NNS) is to find a subset $K\subseteq P$ of size $k$ s.t.\vspace{-.1in} 
$$\min_{p\in P\setminus K} d(p,q) \geq \max_{p\in K} d(p,q)\vspace{-.1in} $$
where $d(p,q)$ is the distance between $p$ and $q$ in $\mathbb{R}^d$.
Intuitively, $K$ contains the top-$k$ points closest to $q$.
Since computing exact $k$-NNS is expensive in high dimensions,
\emph{approximate} nearest neighbor search (ANNS) is widely used in practice.
%With clear context, we omit $k$ and refer to the problem as NNS and ANNS, respectively.

\hide{To measure the quality of the \emph{approximate} $k$-NN search, we use the standard $k@k'$ recall.
% The accuracy of the results from $k$-ANNS is measured by $k@k'$ recall.
Let the set of points $K$ be the true $k$-nearest neighbors of a queried point $q$, 
and $K'$ the set of the approximate nearest neighbors with size $k'$ returned by an ANNS algorithm. 
The $k@k'$ recall of query point $q$ is $r_q=\frac{|K\cap K'|}{|K|}$, and $k@k'$ recall of a query set $Q$ is computed by $r_Q=\sum_{q\in Q}r_q/|Q|$.
Unless specified, we use $k=k'$ in this paper.
To measure efficiency, we use \emph{throughput}. 
If an ANNS algorithm processes a query set $Q$ in running time $t$, the throughput of an ANNS algorithm is $|Q|/t$. 
%For efficiency of ANNS queries, we use \emph{throughput}, which is $|Q|/t$ where $Q$ is the query set
%Denoting the time to answer all queries in $Q$ as $t$, the throughput regarding $Q$ is computed as $|Q|/t$.
}
We use the standard $k@k'$ recall to measure ANNS quality. 
Let $K$ be the set of true $k$-nearest neighbors of a query point $q$,
and $K'$ the set of $k'$ ANNs returned by an ANNS algorithm.
The $k@k'$ recall measures how many of the true $k$-NNs appear in $K'$, 
computed as $r_q=\frac{|K\cap K'|}{|K|}$ for $q$, and $r_Q=\sum_{q\in Q}r_q/|Q|$ for a query set $Q$.
Unless otherwise specified, we use $k=k'$ throughout this paper.
To measure efficiency, we use \emph{throughput}, defined as $|Q|/t$, the number of queries in $Q$ processed in running time $t$.

\subsection{Graph-based Algorithms for ANNS}
%\yan{switch the next two sentences.}
\hide{A non-trivial ANNS algorithm needs to maintain an \emph{ANN index} to accelerate the incoming queries, which is the key distinction between different algorithms.
Such index usually fall into four categories: graphs~\cite{wang2021comprehensive, malkov2020hnsw, dong2011efficient, fu2019nsg, fu2021high, zhang2022hierarchical, munoz2019hcnng, peng2022speed, peng2023iqan, lu2021hvs, harwood2016fanng, iwasaki2016pruned, toussaint1980relative, xu2022proximity, carey2022star, wang2025accelerating}, locality-sensitive hash tables~\cite{mou2017refined, datar2004locality, gionis1999similarity, pham2022falconn, jafari2021survey}, inverted indices~\cite{babenko2014inverted, baranchuk2018revisiting, chen2021spann}, and trees~\cite{raoofy2023overcoming, beygelzimer2006cover, gu2022parallel, yesantharao2021parallel}.
Among them, graph-based ANNS algorithms are particularly effective at achieving high recall, while maintaining efficient throughput. }
\annlib{} focuses on efficient and general implementations of graph-based ANNS, where a graph is constructed to facilitate search. 
Each vertex corresponds to a point in the base set, and edges represent proximity relationships. 
\hide{To control the size of the graph, the degree of each vertex is controlled to be at most $r$, called the \emph{degree bound}. }
To keep the graph size under control, the degree of each vertex is capped at $r$, called the \emph{degree bound}. 
Both the graph construction and the query algorithms build on a common primitive, \emph{beam search}, discussed below.

\hide{The ANNS is usually performed by a heuristic search on the graph along the edges, represented by \bsearch{} and its variants~\cite{?}.
The key to build such a graph is to determine the neighbors of each vertex to reflect the relationship with its nearby points.
Existing algorithms~\cite{?} explore the way that \prune{} the coarsely selected candidates to preserve the most essential ones under the limited \emph{degree bound}.
}

% covering the methods of %% citations below
% bucketing-based algorithms, %% wording
% tree-based algorithms,
% graph-based algorithms,
% and space compression techniques
\hide{
\begin{itemize}
\item FAISS (IVF-based)
\item ParlayANN (graph-based)
\item hnswlib (nmslib) (graph-based)
\end{itemize}
}

\myparagraph{Beam Search.} 
A beam search traverses a graph $G=(V,E)$ from a starting point $s\in V$ towards the query point $q$. %The high-level idea is to 
As shown in~\cref{algo:bsearch}, it maintains a candidate queue $\mathcal{C}$ with the capacity of beam width $L$, which initially only contains $s$ and keeps updating iteratively.
At each iteration, the algorithm retrieves the nearest point to $q$ among the candidates in $\mathcal{C}$, and visits its unvisited neighbors (line \ref{line:bsearch:get}). 
% If the newly visited point has shorter distance than any of those in $\mathcal{C}$, or $\mathcal{C}$ is not full, the point is pushed into $\mathcal{C}$, and the one with longest distance, if $\mathcal{C}$'s size exceeds the beam size, is popped.
\hide{Upon newly visiting a point $u$, $u$ is added to $\mathcal{C}$.}
Each newly visited point $u$ is added to $\mathcal{C}$.
If the size of $\mathcal{C}$ exceeds $L$, only the $L$ closest points to $q$ are kept (line \ref{line:bsearch:evict}).
\begin{algorithm2e}
	\caption{BeamSearch($G,s,q,L$)}
	\label{algo:bsearch}
\fontsize{9pt}{9pt}\selectfont
	% \SetKwBlock{ParDo}{do in parallel}{end}
	% \SetKwBlock{ParFor}{parallel for}{end}
    \DontPrintSemicolon
	\SetAlgoLined
	\KwIn{Graph $G=(V,E)$, starting point $s$, query vertex $q$, and beam width $L$}
	\KwOut{the $L$ closest vertices to $q$ visited during search}
	$\mathcal{V} \gets \emptyset$; $\mathcal{C} \gets \{s\}$ \;
	\While{$\mathcal{C}\setminus\mathcal{V} \neq \emptyset$}{
		$ u \gets \arg\min_{p\in\mathcal{C}\setminus\mathcal{V}}d(p,q) $\;\label{line:bsearch:get}
		$ \mathcal{V} \gets \mathcal{V} \cup \{u\} $\;
		$ \mathcal{C} \gets \mathcal{C} \cup \{v|(u,v)\in E\} $\;
		\lIf{$|\mathcal{C}|>L$}{
			keep $L$ closest vertices to $q$ \label{line:bsearch:evict}
		}
	}
	\Return{$\mathcal{V}$}
\end{algorithm2e}
% It maintains a beam $\mathcal{C}$ with size at most $L$ as the candidates currently closest to $q$. $\mathcal{C}$ is initialized with the starting point and updated iteratively.
% In each iteration, the algorithm retrieves the vertex that is unvisited and closet to $q$, and visits all its unvisited neighbors. Upon visiting a vertex $u$, the distance $d(u,q)$ is computed, and $u$ is added to the beam. If the size of $\mathcal{C}$ exceeds $L$, only the $L$ closest vertices to $q$ are kept.

\begin{figure}[t]
  \centering
  \includegraphics[width=\linewidth]{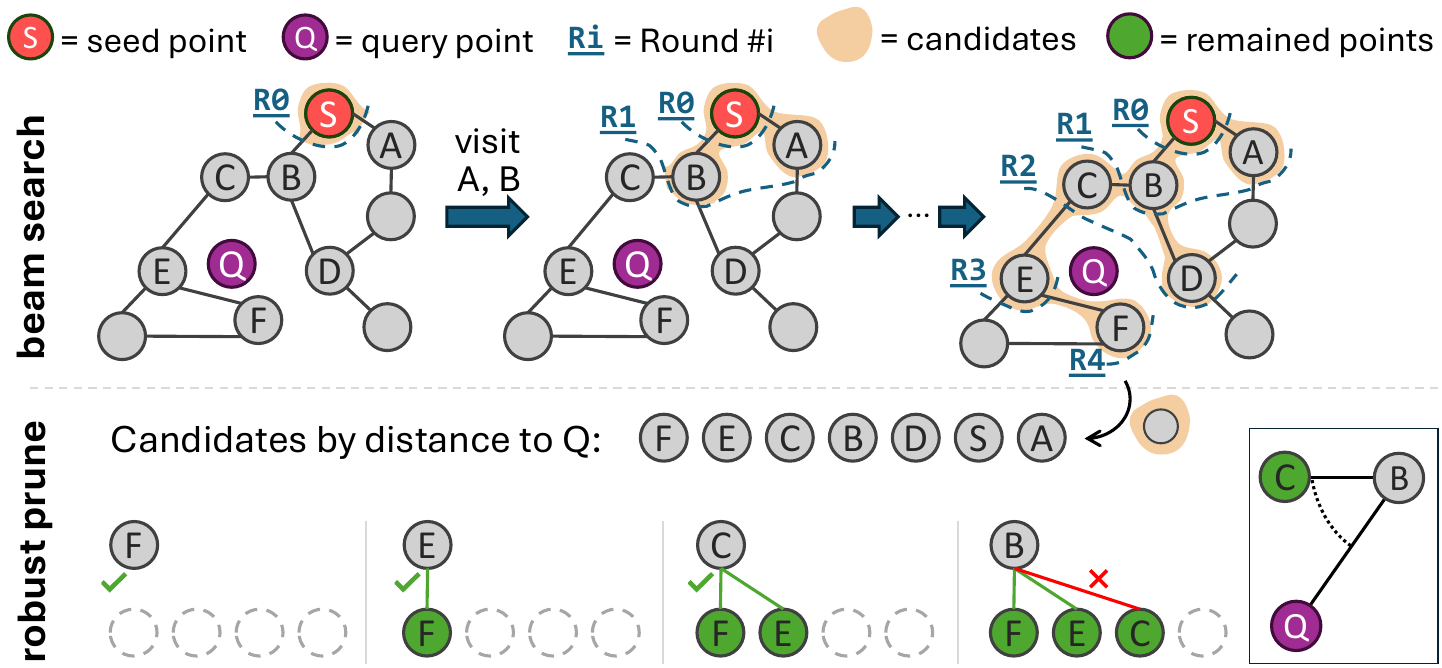}
  \caption{Examples of the beam search and the robust prune.}
  \label{fig:search-and-prune}
\end{figure}

\hide{\Cref{fig:search-and-prune} gives an example of the beam search starting from $S$. Suppose the beam width $L=2$.
At the first round, $S$ checks its neighbors, $A$ and $B$, and pushes them into $\mathcal{C}$. No vertex is evicted from $\mathcal{C}$.
Next, the closest point, $B$, expends to $C$ and $D$ and pushes them to $\mathcal{C}$, making $A$ evicted. $S$ is skipped as it has been visited. 
In the following rounds, $C$ expends to $E$, and $E$ expends to $F$.
The algorithm eventually returns the visited points, as highlighted in the figure.}
% searching from $S$ towards $Q$.

\myparagraph{Pruning.} 
\hide{As the graph is constructed, new vertices may be inserted, introducing new edges to existing vertices.}
During graph construction, newly inserted vertices may add new edges to existing vertices.
In this case, a \emph{pruning} algorithm is typically adopted to refine the neighborhoods of vertices. 
\hide{The pruning algorithm selects \(r\) (called the \emph{degree bound}) neighbors for a vertex \(u \in V\) from candidates \(\mathcal{C}\), with \(|\mathcal{C}| \geq r\). }
Given a candidate set $\mathcal{C}$ with $|\mathcal{C}|\geq r$, the pruning algorithm selects $r$ neighbors for a vertex $u\in V$.
We present the algorithm in~\cref{algo:prune}. 
\hide{The algorithm determines whether each point in \(\mathcal{C}\) is pruned in order of distance to \(u\) (lines~\ref{line:prune:sort}--\ref{line:prune:get}). }
The algorithm determines whether to prune each point in $\mathcal{C}$ in ascending order of distance to $u$ (lines~\ref{line:prune:sort}--\ref{line:prune:get}).
Each pruning decision is made by a predicate (line~\ref{line:prune:pred}), based on the distances to \(u\) and to the already-selected neighbors.% points that are preserved.

A simple strategy~\cite{malkov2020hnsw}, known as \textit{re-ranking}, considers only the distance to $u$ and selects the $r$ nearest points. 
It is typically used to determine the final answers to a query.
In this case, \textit{pred} always returns \textit{false}. 
\hide{Another commonly-used heuristic, \textit{robustPrune}, optimize the neighbor distribution thus improving the index quality.}
Another commonly used heuristic, \textit{robustPrune}~\cite{subramanya2019diskann}, optimizes the neighbor distribution to improve the index quality.
In \textit{robustPrune}, the current point $p^*$ is pruned if it is closer to any already-selected point $p'$ than to $u$, i.e., \textit{pred} holds when $d(p^*,p')<d(p^*,u)$.
The intuition is that, since $p'$ is preserved and $p^*$ is closer to $p'$, the search is likely to reach $p^*$ through $p'$, making $p^*$ redundant.
% There exist other variants for the predicate~\cite{gollapudi2023filtered,fu2019nsg}.
Other predicate variants exist, such as ~\cite{gollapudi2023filtered,fu2019nsg}.
% neighbors of the target point $u$ from $m$ candidates ($r<m$) to make the selected ones locate near $u$ in varying directions for easy navigation to $u$. 
\begin{algorithm2e}
	\caption{prune($u, \mathcal{C}, r, \textit{pred}$)}
	\label{algo:prune}%\footnotesize
    \fontsize{9pt}{9pt}\selectfont
	% \SetKwBlock{ParDo}{do in parallel}{end}
	% \SetKwBlock{ParFor}{parallel for}{end}
    \DontPrintSemicolon
	\SetAlgoLined
	\KwIn{Point $u$, candidate set $\mathcal{C}$, degree bound $r$, predicate \textit{pred}}
	\KwOut{Selected neighbors $\Nu$}
	sort $\mathcal{C}$ by distance to $u$ in an ascending order\;\label{line:prune:sort}
	$\Nu \gets \emptyset$ \;
	% \While{$\mathcal{C} \neq \emptyset$}{
	% 	$ p^* \gets \arg\min_{p'\in\mathcal{C}}d(u,p') $\;\label{line:prune:get}

	\For{$p^* \in \mathcal{C}$ (in sorted order)}{\label{line:prune:get}
		$ \Nu \gets \Nu \cup \{p^*\} $\;
		\For{\upshape $p' \in \Nu$}{
			\lIf{\upshape \emph{pred}($u,p',p^*$)}{\label{line:prune:pred}
				remove $p^*$ from $\Nu$ and \textbf{break}
			}
		}
		\lIf{$\left|\Nu\right|=r$}{
			\textbf{break}
		}
	}
	\Return{$\Nu$}
\end{algorithm2e}
\Cref{fig:search-and-prune} illustrates this process for determining the neighbors of $Q$. 
First, a beam search starting from seed $S$ visits its neighbors $A$ and $B$, which forms the candidate set. 
Next, it picks the current nearest point $B$ and visits $B$'s neighbors $C$ and $D$, and so on.
The beam search ends at round~4 and outputs the candidates sorted by their distances to $Q$.
\textit{robustPrune} then selects the final neighbors from the candidates by degree bound $r=4$,
where $F$, $E$, and $C$ are preserved, while $B$ is pruned since it fails the predicate against $C$. 
In the next section, we will introduce our techniques, which encapsulate the algorithms reviewed above and new data structures under a unified abstraction, enabling adaptation to more scenarios with minimal effort.

% \section{Facilitate the Development}
\section{The ANNLib Design}
\label{sec:design}
%% 3.1
%% ADD subsubsection
%% 1. insert + query
%% 2. f_nbhs/f_dist design (put closer to the caller)
%% 3.2
%% section
\subsection{Overview}
\label{sec:design:overview}
\hide{
The design principle of \annlib{} is to decouple the components in an ANN system so that each part can be swapped in a plug-and-play manner and developed individually.
Although decoupling is a classic concept in software design, \annlib{} makes strong efforts to achieve both programming flexibility and performance efficiency.
The challenge is how to satisfy varying customization from not only end users but also developers at different levels, and how to improve performance for complicated and even unknown applications.
}
%The design principle of \annlib{} is to decouple the components in the ANN system so that each part is plug-and-play and can be developed individually.
%Although decoupling is a classic concept in software design,
%achieving it without sacrificing flexibility or performance is challenging. 
%Specifically, \annlib{} must meet the customization needs of both end users and developers at different levels,
%while sustaining performance for complex and even unknown applications.

\annlib{} is designed to decouple ANNS components into modular, plug-and-play units. 
However, achieving such decoupling without compromising flexibility or performance is challenging, especially when supporting multi-level customization (for both developers and users) across complex applications. 
To address this, we partition \annlib{} into two core components: \emph{parallel algorithms} and \emph{parallel data structures}. 
The algorithm part includes \emph{base algorithms} for index construction (integrating Vamana~\cite{subramanya2019diskann} and HNSW~\cite{malkov2020hnsw}, while allowing user extensions)
and \emph{algorithm modules} for advanced functionalities (e.g., filtered queries and updates). 
The \emph{data structure} manages the underlying graph, providing unified interfaces for reading and modifying it.

%To this end, we design \annlib{} as two components: the algorithm and the data structure. 
%We further divide the algorithm component into two parts: the \emph{base algorithms} and \emph{algorithm modules}.
%The base algorithms specify how the graph index is built.
%For example, \annlib{} integrates Vamana~\cite{chen2021spann,subramanya2019diskann}, HNSW~\cite{malkov2020hnsw}, and HCNNG~\cite{munoz2019hcnng} as base algorithms, and supports new algorithms added by users.
%The algorithm modules provide additional functionality, such as historical queries, filtered queries, and dynamic insertions and deletions.
%The \emph{data structure} component maintains the underlying graph. 
%Specifically, it provides interfaces for algorithms (and users) to read and modify the graph container.

\Cref{code:use-hnsw} shows how to specify which components to use and apply customization in actual code.
On line~\ref{line:hnsw-algo}, the \emph{HNSW} algorithm is selected, along with other components described in \textit{desc}.
The \textit{desc} specifies the graph container (line \ref{line:graph-t}) and other user-defined types, such as the point type and distance function.
To switch to another algorithm such as \vamana{}, the user simply changes the algorithm name on line~\ref{line:vamana-algo}, leaving the other parts still.
% Similarly, one can make a single change on line \ref{line:graph-t-pam} to alter the index type and obtain functional-update capability from a PAM-based graph~\cite{Sun2018PAM}.
Similarly, graph container can be changed solely on line \ref{line:graph-t}.

\begin{lstlisting}[
	caption={Example of invoking the pre-built HNSW algorithm, using the adj\_seq graph and the direct mapping.},
	label={code:use-hnsw},
	language=C++,
	emph={HNSW,desc},
	escapechar=@
]
struct desc{
	using point_t = ...;@\label{line:point-t}@
	using graph_t = ANN::graph::adj_seq;@\label{line:graph-t}@
	auto dist_func = ...; @\label{line:dist-func-t}@
	... };
std::vector<point_t> ps = ...;
ANN::algo::HNSW<desc> index;@\label{line:hnsw-algo}@
// Alternatively, ANN::vamana<desc> index;@\label{line:vamana-algo}@
index.insert(ps); // batch insertion
index.search(q, 10); // 10-NN search regarding q
\end{lstlisting}
%// ANN::algo::vamana<desc> index;@\label{line:vamana-algo}@

\hide{Furthermore, in need of customizing lower-level logic,
one can partially overload existing building blocks as long as they
follow the interface definition, which is detailed in the rest of this section.}
Furthermore, if lower-level logic needs customization,
one can partially redefine existing building blocks,
as long as they follow the interface definition (see details below).

\Cref{code:example-filter} shows how to support filtered search by customizing the predicate on top of \vamana{}.
\hide{
At line \ref{line:vamana-ins}, the original \vamana{} in \annlib{} invokes the regular pruning algorithm for index build.
To develop \fvamana{}, a feasible approach is to inherit \vamana{} and modify the insertion method, passing a custom predicate that considers labels.
}
On line \ref{line:vamana-ins}, the original \vamana{} invokes the regular pruning algorithm during index construction.
To develop \fvamana{}, one can inherit from \vamana{} and redefine \texttt{insert} to pass a custom predicate that considers labels.

\begin{lstlisting}[
	caption={Example of implementing filtered search.},
	label={code:example-filter},
	language=C++,
	emph={pred\_reg,pred\_filter},
	escapechar=@
]
// in vamana::insert
auto nbhs = prune(cands, pred_reg, ...);@\label{line:vamana-ins}@

// in filtered_vamana::insert @\label{line:vamana_ins}@
auto pred_filter = [&](...){/*consider labels*/};
auto nbhs = prune(cands, pred_filter, ...);


\end{lstlisting}

\subsection{Algorithms: Common Logic Abstraction}
\hide{
The underlying algorithm is the core logic component that defines how to build and search an ANN index.  
\annlib{} eliminates unnecessary effort in developing ANN algorithms by abstracting common logic.  
The motivation is that, if the unique logic in an ANN algorithm is separated and wrapped as parameters to general functions, developers can focus solely on implementing those unique parts.
}
\hide{
We explain the design details starting from the algorithm component, which is 
the core logic of defining how to build and search an ANN index.
\annlib{} contributes to algorithm development by minimizing the required effort to the only necessary amount.
Users only pay for the unique part of their ANN algorithm, and \annlib{} handles the rest by abstracting the common logic.  
}
The algorithm component is the core of ANNS index construction and query,
and where most development effort goes.
% \annlib{} minimizes this effort by letting users implement only the part unique to their algorithm,
% and abstracts away the common logic.
\annlib{} minimizes this effort by letting users implement only the part unique to their algorithm, while abstracting away the common logic.
\begin{lstlisting}[
	language=C++,
	caption={A typical insertion procedure. },
	label={code:general-insert},
	escapechar=@,
]
auto collect(f_nbhs, f_dist, eps, r)
auto prune(cand, rsize, f_dist, pred)
@\textrm{\textit{\color{commcol}// Insert \texttt{points} along with their neighbors into the graph}}@
void insert(auto points){
	auto new_nbhs[];
	(parallel-)for(p : points){@\label{line:pfor}@
		new_nbhs[p] = comp_nbhs(p, max_deg);
		// auto cands = collect(p);@\label{line:bsearch}@
		// new_nbhs[p] = prune(cands, max_deg);@\label{line:prune}@
	}
	graph.set_edges(new_nbhs); }
\end{lstlisting}
%%@\textrm{\footnotesize\ }@
%	linebackgroundcolor={
%		\ifnum\value{lstnumber}=4
%			\color{orange!30}
%		\fi
%	}

% \subsubsection{The (Batch) Insertion Algorithm as an Example of Algorithmic Abstraction}
\subsubsection{Batch Insertion as an Example of Algorithmic Abstraction}
We present the abstract insertion procedure as an example --- an important subroutine for various ANNS algorithms~\cite{subramanya2019diskann,malkov2020hnsw,munoz2019hcnng,zhang2022hierarchical,chen2018sptag} that build indexes via incremental insertion.
\Cref{code:general-insert} shows a typical insertion workflow, with  the key logic highlighted inside the for-loop: determining the neighbors of the newly inserted point $p$. 
%\annlib{} uses the prefix-doubling technique  ParlayANN~\cite{manohar2024parlayann} to handle parallelism, which insert points in increasing sizes of batches in a race-free manner. 
We also support parallel batch computation (using parallel for-loops), as many applications require it.
\hide{
Many applications require the ability to insert batches of points in parallel.
To facilitate such cases, the \texttt{insert} function also supports a parallel for-loop to handle all points in parallel. 
}

Computing a point's neighbors usually takes two steps:
collecting candidates (function \texttt{collect}, line \ref{line:bsearch}) and selecting final neighbors (function \texttt{prune}, line \ref{line:prune}). 
%When optional parallelism is enabled on line \ref{line:pfor}, 
%the lost relationships among intra-batch points can be handled by the prefix-doubling technique from ParlayANN~\cite{manohar2024parlayann}. %\yan{what does this sentence mean?}
These two steps can be solved by combinations of primitives \bsearch{} and \prune{}, introduced in \cref{sec:background}. 
For example, HNSW~\cite{malkov2020hnsw} performs beam search to $p$ at each layer to collect candidates and then applies either a simple sort or a heuristic to prune neighbors. 
HCNNG~\cite{munoz2019hcnng} uses 3-MST at leaf nodes to compute candidate edges in a single tree, then merges them across trees with pruning strategies.
To express all of these in a unified interface,
we encapsulate \bsearch{} and \prune{} and adapt them across multiple ANNS algorithms and variants (detailed in \cref{sec:app}).
Such extensibility is achieved by adhering to the actual data requirements of \bsearch{} and \prune{}: across algorithms, only distance computation and the neighbor list are needed. 
We therefore expose them through two parameters,
\fnbhs{} and \fdist{}, on the \texttt{collect} and \texttt{prune} interfaces.
\fnbhs{($u$)} $\to \{v:(u,v)\in E\}$ is a callable object % ALTER: closure
that returns the reachable neighbors of $u$.
\fdist{($u$)}$: u\to d(u,q)$ is also callable and returns the distance between $u$ and the query point;
an additional overload \fdist{($u,v$)} returns the distance between $u$ and $v$. 
\pred{($p',p^*$)} $\to$ \textit{bool} accepts a candidate $p^*\in$ \texttt{cand} and a selected neighbor $p'$,
deciding whether $p^*$ should be selected with respect to $p'$.
We now present how these objects are defined for regular search.

\begin{lstlisting}[
	language=C++,
	caption={Examples of \fnbhs{}, \fdist{}, and \pred{}},
	label={code:example},
	emph={f\_nbhs,f\_dist,pred}
]
f_nbhs = [g](u){return g.get_edges(u);};
f_dist = [g,q](u){return g.dist(u,q);};
pred = [](pp,ps){return false;};
\end{lstlisting}

% To match the generic use of these algorithms, the parameters are carefully designed, based on the problem essence, to best fit the logic of the developers' new algorithms. The beam search accepts $f\_nbhs$, $f\_dist$, $eps$, $ef$, and $opts$. $f\_nbhs$ is a function that accepts a single parameter $u$ and returns the edge list of the node $u$. $f\_dist$ is also a function that accepts a single node id $v$ and returns the distance to the query target $d(v,q)$.
Instead of explicitly passing the query point $q$ as a parameter~\cite{manohar2024parlayann,subramanya2019diskann,hnswlib2019hnswlib}, 
our design embeds $q$ inside \fdist{} to handle the cases where $q$ is virtual, as in filtered search.
\hide{
By encapsulating $g$'s information separately in \fnbhs{} and \fdist{},
\annlib{} does not require
the vertex for neighbor traversal and for distance computation come from the same graph
and supports multi-layer-graph algorithms such as HNSW, 
where only the bottom layer stores the coordinates and each layer has its own connections.
}
% \edit{
By encapsulating $g$'s information separately in \fnbhs{} and \fdist{},
\annlib{} does not require the vertices used for neighbor traversal and for distance computation to come from the same graph,
supporting multi-layer algorithms such as HNSW, where only the bottom layer stores coordinates.
% }
Besides, the indirection of \fnbhs{} makes it easy to manipulate the neighbor list for extended applications.
For example, deletion is supported by filtering out tombstone-marked neighbors on the fly.
The remaining parameters for \texttt{collect} and \texttt{prune} include the number of returned edges (degree bound) \texttt{r}, the entry points \texttt{eps}, and the candidates \texttt{cand}.
With \annlib{}, developers can reuse this shared logic to implement and optimize multiple algorithms and strategies by simply plugging in these parameters and helper functions.

\subsubsection{Base Algorithms and Algorithm Modules}
\annlib{} supports a variety of \emph{base algorithms} that can be easily plugged in, including built-in \vamana{}~\cite{subramanya2019diskann}, HNSW~\cite{malkov2020hnsw} and HCNNG~\cite{munoz2019hcnng}.
Many are carefully selected from state-of-the-art solutions with proven efficiency.
They can all be combined with additional functionalities, which we call \emph{algorithm modules}, including construction, filtered search, deletions, and snapshots.
For example, our construction module builds on the prefix-doubling scheme~\cite{manohar2024parlayann}, which calls the \texttt{insert} algorithm in increasing batch sizes, enabling parallel insertion of all points in a race-free manner.
We further introduce other modules in \cref{sec:app}. 
\subsection{Data Structures: Graph Containers}
\label{sec:design:graph}
\hide{
The graph container's role is to let ANN developers easily choose a high-performance data structure that best fits their target scenario for storing the graph index.
For example, when only insertions are needed, the nested-array container---vertices in an array and each vertex's incident edges in a nested array---is suitable due to its good locality and contiguous-memory access.
In contrast, highly dynamic workloads with frequent copies and snapshots benefit from a tree-embedded container that handles updates efficiently.
\annlib{} supports both behind a uniform interface.
Below, we first present the interfaces for integrating arbitrary containers, then show how the cases above use them.
}
% \edit{
The graph container lets developers choose the data structure that best fits their scenario.
For example, a nested array suits static, insertion-only workloads for its good locality,
while a tree-embedded container suits highly dynamic, snapshotted ones.
\annlib{} supports both behind a uniform interface.
Below, we present the container interfaces and how the cases above use them.
% }
\begin{lstlisting}[
	language=C++,
	caption={Interfaces of a graph container},
	label={code:graph-api},
	emph={vid\_t,vertex\_ptr,agent},
	emphstyle=\bfseries
]
using vid_t;
using struct vertex_ptr;
// return an extended-range type
agent get_edges(vid_t/vertex_ptr u);
void set_edges(seq<(vid, edges)>);
// vertex operations
vertex_ptr get_vertex(vid_t u);
void add_vertices(seq vertices);
void remove_vertices(seq vids);
\end{lstlisting}
% void add_vertex(nid_t/vertex_ptr u, auto ext);
% void remove_vertex(nid_t/vertex_ptr u);
% size_t num_vertices();	// num of total vertices
% size_t num_edges();		// num of total edges
% void iter_each(auto f);//apply f to all vtx (seq)
% void for_each(auto f); //apply f to all vtx (par)

\Cref{code:graph-api} lists the minimal interfaces to support diverse ANNS algorithms.  
To enable batch operations, \texttt{add\_vertices} and \texttt{remove\_vertices} accept multiple vertices,
and \texttt{set\_edges} accepts a sequence of vertex--edges pairs% indicating which vertex to update and its associated edges.  
, each indicating a vertex to update and its associated edges.
The interfaces require only basic graph operations by design for ease of use, but
the challenge is to support complex use cases efficiently across varying containers.  
\annlib{} addresses this by separating container-specific details from the interface definition and
encapsulating them into an \agent{}.

% Theses operations can be simply forward to underlying graph and work with 
% To support parallel updates in batch style, a concept of edge agent is introduced. An edge agent is a brokerage of an edge list. When the user gain an edge agent, the ownership of the corresponding edge list is borrowed (late transferred) to the agent. If the agent is not modified, the ownership keeps intact (unchanged) in the end. The read operations that access to the agent is equivalent to as to the original edge list. However, the write (assignment) operation automatically transfer the ownership to the agent, and the changes are conceptually store in the agent. When all updates have been done, the agent have to be committed to eventually apply the changes.
\subsubsection{Edge Agent} 
\hide{We introduce \agent{}, an intermediate structure that optionally enables graph-specific optimizations.
The key idea is to push technical, container-specific details down into \agent{}---implemented by the graph container for maximal performance---while keeping the upper interface used by algorithm developers uniform and simple.
}
The \agent{} is an intermediate structure that lets each container apply its own graph-specific optimizations.
The key idea is to push container-specific details down into the \agent{}, which each container implements for maximal performance,
so that algorithm developers see a uniform, simple interface.
The \agent{} for a vertex $u$ is produced by \texttt{get\_edges($u$)}.
It acts as a view over $u$'s edges while also holding temporary ownership of them during its lifetime.
Below, we illustrate the use of \agent{} with examples that insert back-edges across graph containers.

\subsubsection{Back Edge Insertion} 
Back-edge insertion is a subroutine of batch insertion in many algorithms~\cite{manohar2024parlayann}.
Given a vertex $u$, an algorithm attempts to add a set of vertices $\mathit{es}$ as new neighbors of $u$.
In most cases, this requires jointly evaluating $\mathit{es}$ and the existing neighbors of $u$,
and re-selecting at most $r$ of them as $u$'s new neighbors, where $r$ is the degree bound. 
The primary steps are shown in \cref{code:back-insert}.
\begin{lstlisting}[
	language=C++,
	caption={The primary steps of back edge insertion.},
	label={code:back-insert},
	emph={get\_edges,set\_edges,agent,parallel_for},
	emphstyle=\bfseries,
	escapechar=@
]
@\textrm{\textit{// For each forward edge ($u,v$), insert ($v,u$) to the graph in batch}}@
updates = rev_and_semisort(forward_edges)
auto edge_agents[];
parallel_for(u, es : updates){
	agent = graph.get_edges(u);@\label{line:get-edges}@
	old = move_to_seq(agent);@\label{line:move-to-seq}@
	agent = prune(old + es);@\label{line:write-back-agent}@
	edge_agents = move(agent);  }@\label{line:set-agent}@
graph.set_edges(edge_agents);@\label{line:set-edges}@
\end{lstlisting}
Here, \texttt{updates} is the batch containing vertices and their new edges to be added.
On lines~\ref{line:get-edges}--\ref{line:write-back-agent}, for each vertex $u$,
the existing edges are retrieved, merged with the new edges,
and the final neighbors are written back via the agent.
The same code snippet works across both nested-array and tree-embedded graph containers because each provides its own \agent{} backend.

In the following sections, we introduce the built-in containers in \annlib{}, each backed by a different data structure.

\subsubsection{Nested array}\label{design:nested-array}
A nested array stores each vertex's edges in an array, and the agent behaves like a reference to that array. %They are usually best suited for static settings where the graph is not updated. 
During updates, since $u$ will replace its edges with a new array, there is no need to preserve the array referenced by agent; on line \ref{line:move-to-seq}, it is trivially moved into \texttt{old} because the agent owns it.
Moreover, the agent directly forwards the write on line \ref{line:write-back-agent}, completing the insertion early; consequently, lines \ref{line:set-agent} and \ref{line:set-edges} are no-ops.
The nested array provides good locality and space efficiency for accessing the graph, but is expensive to update.

\subsubsection{Functional Tree}
To support snapshottable graph, we use functional tree. 
Using functional trees for dynamic graph with snapshots was first proposed in~\cite{dhulipala2019ctree}. 
We use a follow-up work called CPAM~\cite{dhulipala2022pac}.
In general, functional trees use copy-on-write (CoW) upon updates, thus preserving their previous version. 
A graph is represented by a two-level nested tree. 
The outer tree, or the \emph{vertex tree}, has each entry as a vertex, associated with an \emph{inner tree}, which stores all its neighbors. 
Both levels are functional and are snapshottable. 
In this case, the agent holds the tree's root plus an internal array to stage updated edges.
Since the edge tree nodes can be shared due to CoW, \texttt{move\_to\_seq} on line~\ref{line:move-to-seq} effectively flattens the tree into an array.
The assignment to \texttt{agent} on line \ref{line:write-back-agent} writes the final neighbors into the internal array, rather than the tree itself, to avoid immediate mutations that would ripple into the outer vertex tree and cause data races.
Agents are then gathered (line \ref{line:set-agent}) and submitted as a batch (line \ref{line:set-edges}) to apply changes top-down. 
Using CPAM, one can effectively run historical queries on a previous snapshot. 

\subsubsection{The Chrono Prefix Array}\label{design:chrono-prefix-array}
Although CPAM provides a direct solution for historical ANNS queries, it is not always efficient. 
Since the degree bound is usually small (30--100) relative to the block size (typically 64), each inner tree holds only a few nodes,
and at such small size the tree's structural overhead outweighs its benefits. 
We design a new data structure for ANNS graphs with historical queries, called the \emph{chrono prefix array}. %chained prefix array??
For each vertex, it maintains a linked list of its edge versions over time, 
where edges are stored in a space-saving structure, \parray{}.
A \parray{} compresses chronologically consecutive edge versions together by sharing their prefix. 
It consists of a reference-counted buffer $R$ and multiple cursors $c_1\dots c_k$, where $k$ is the number of versions.
Each combination of a cursor and the buffer $(c_i,R)$ represents an effective version of edges in $R_{1\dots c_i}$.
When new edges $E$ arrive, it computes the increment $M = E \backslash R_{1\dots c_k}$ and appends $M$ to $R$.
A new version $(c_{k+1=|R|}, R)$ is thereby created for the changes of $E$.
Note that $R_{1\dots c_{k+1}}$ may contain more edges than $E$, since points appearing in $R_{1\dots c_k}$ but not in $E$ are also included.
We argue that it is usually safe to contain these extra edges
because such deviations are small in practice and extra edges in beam search do not impair the query recall; they just slightly increase the search time.
% In one hand, such deviation is small since $E$ differs a little from the last version $R_{1\dots c_k}$ during the index building.
% In the other hand, in the beam search, extra edges do not impair the query recall but only slightly increase the search time.
$R$ has a size limit.
Once the number of new edges exceeds the limit, a new array is allocated.

\begin{figure}[h]
  \centering
  \includegraphics[width=\linewidth]{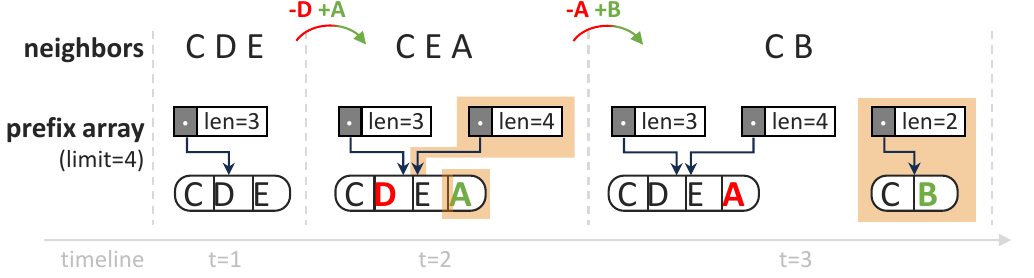}
  \caption{Examples of preserving all changes on a chrono list. \label{fig:chrono}}
\end{figure}

\hide{
\Cref{fig:chrono} provides an example to explain how it works.
At the moment $t=1$, the initial edge version contains $C,D,E$, stored in the buffer with a cursor of 3.
When $t=2$, the edges are changed to $C,E,A$. The \parray{} computes the increment $A$ and appends it to the end with a new cursor of 4.
A new edge version is $C,D,E,A$ where $D$ is the harmless extra edge.
At $t=3$, edges are changed to $C,B$. Since the previous array has reached its limit of 4, a new one is created to contain edges.
}
% \edit{
\Cref{fig:chrono} shows an example:
a new version appends only the increment (e.g., $A$ at $t{=}2$) and may retain a harmless extra edge (e.g., $D$).
Once the buffer hits its size limit, a fresh array is allocated (at $t{=}3$).
% }

\section{Applications and Implementations}\label{sec:app}
%In this section, we demonstrate the applications of \annlib{}, regarding how developers can fast implement batch insertion, batch deletion, filtered search, and snapshots on the top of \annlib{} with only necessary effort of modification. These applications will be further evaluated in \cref{sec:eval}.

This section presents five applications built on \annlib{}. The first four demonstrate how developers can easily implement advanced features with minimal modification. The fifth is a case study of how re-implementing two existing range-filtered ANN systems on \annlib{} fixes an engineering bottleneck in each, without needing to touch other designs in the original system. These case studies are evaluated in \cref{sec:eval}.

\subsection{Parallel Batch Insertion on \vamana{}}\label{app:bat-ins}
%\annlib{} provides the parallel batch insertion of \vamana{} built upon the general insertion steps in \cref{code:general-insert}, and fills in the \vamana{}-specific logic to compute neighbors. 
We use the parallel batch insertion of \vamana{} as an example of how algorithm-specific logic can be plugged into \annlib{}.
%the HNSW version is given in the supplemental material. 
For each newly inserted point $p$, it performs a \bsearch{} on the existing graph targeting $p$, then applies \robustprune{} to the search results, using \fnbhs{} at line \ref{line:ins-fnbhs-end} and \fdist{} at line \ref{line:ins-fdist-end}.
The back edges of \vamana{} are then inserted following the technique of ParlayANN~\cite{manohar2024parlayann}. 
The \texttt{group\_by} operation is customized to a parallel implementation without changing the invocation at line \ref{line:ins-group}, due to the native parallel design. The \texttt{nbh\_rev[$\cdot$]} is an array of \agent{}s, inserted in the same form as the forward edges.
% based on techniques of ParlayANN~\cite{manohar2024parlayann}for its good performance and scalability. Additionally, it implements batch deletion by marking the tombstone and recomputing the neighbors at the watermark, detailed in \cref{app:bat-del}. The support of such batch operations provides efficient and deterministic parallelism, and the users are allowed to modified the parts of their interest to implement the desired functionalities while still benefiting from the advantages of batch parallelism, without needs of writing the whole logic.

\begin{lstlisting}[
	language=C++,
	caption={\vamana{} Batch Insertion},
	morekeywords={foreach},
	escapechar=@
]
auto get_f_nbhs(){
	return [&](nid_t u){ return g.get_edges(u); }; }@\label{line:ins-fnbhs-end}@
auto get_f_dist(nid_t v){
	return [&](nid_t u){ @\label{line:ins-fdist-begin}@
      return dist(g.get_node(u), g.get_node(v)); };@\label{line:ins-fdist-end}@
}
void insert(auto pts){
	auto nbh_fwd[];
	foreach(p : pts){
		auto cands = beamSearch(p,get_f_nbhs(),get_f_dist(p));
		nbh_fwd[p] = prune(cands, max_deg, get_f_nbhs());  }
	set_edges(nbh_fwd);
	auto nbh_rev[];
	auto edge_rev = {(v,u) | (u,v) in nbh_fwd};
	auto groups = group_by(edge_rev);@\label{line:ins-group}@
	foreach([v, es] : groups){
		nbh_rev[v] = prune(es + get_edges(v), ...); 	}
	set_edges(nbh_rev); }
\end{lstlisting}
\hide{
	linebackgroundcolor={
		\ifnum\value{lstnumber}=2
			\color{orange!30}
		\fi
		\ifnum\value{lstnumber}=5
			\color{orange!30}
		\fi
		\ifnum\value{lstnumber}=8
			\color{orange!30}
		\fi
		\ifnum\value{lstnumber}=15
			\color{orange!30}
		\fi
		\ifnum\value{lstnumber}=18
			\color{orange!30}
		\fi
		\ifnum\value{lstnumber}>21
			\ifnum\value{lstnumber}<29
				\color{orange!30}
			\fi
		\fi
	}
}

\subsection{Batch Deletion}\label{app:bat-del}
As mentioned, while most ANNS algorithms naturally support insertion, deletion is more challenging since removed vertices and edges may disconnect the graph. 
\annlib{} implements the dynamic setting based on the deletion algorithm in FreshDiskANN~\cite{singh2021fresh}, consisting of two steps: 1) mark deleted points and 2) consolidate marked neighbors with edge repairing. We deem deletion marks as a label and pass in a filter to skip marked points during search. To perform  consolidation, we use \annlib{}'s helper to traverse the points in parallel and update neighbors through \agent{}s, reusing the same code path.
%For page limit, we present the algorithm and pseudocode in the supplemental material. 

\subsection{Filtered Search}\label{app:flt-search}
Let $\mathcal{F}$ denote the universe of labels.
Given an input set $P$ where each point is associated with \emph{labels} defined by $F: P \to 2^{\mathcal{F}}$,
the filtered ANNS aims to find \knn{s} of a query point $q$ with specified labels $F_q$ that satisfy $F_q \subseteq F_u$ for each $u\in P$ in the result.
Filtered ANNS is inherently difficult, especially on skewed distributions of labels.
\annlib{} provides it as a built-in block based on the \svamana{} and \fvamana{} algorithms of Filtered-DiskANN~\cite{gollapudi2023filtered}. \svamana{} builds a regular \vamana{} graph per label and merges the sub-graphs by \fprune{}, so it inherits the insertion implementation for the first step and needs only a new \texttt{merge} function for the second. \fvamana{} instead relaxes the pruning condition, which we express as a custom \textit{f\_nbhs}. Search for either reuses the regular \textit{beam search} with an \textit{f\_nbhs} that filters neighbors by label. We further parallelize the per-label merge, which the original algorithm performs sequentially.
We present the algorithms and pseudocode in Appendix~\ref{app:flt-search-detail}.

\subsection{Snapshots}\label{app:snapshot}
We develop the snapshot functionality, which saves the whole ANN index for future use.
This is achieved by making the graph container snapshottable, without changing the algorithms.
As demonstrated in \cref{code:snapshot}, an ANN snapshot is conceptually a historical copy,
where the underlying data movements are managed by space-efficient data structures.
\begin{lstlisting}[
	language=C++,
	caption={Create a snapshot},
	label={code:snapshot},
	morekeywords={},
	escapechar=@
]
snapshot1 = (index, tick++);
index.insert(points);
snapshot2 = (index, tick++);
\end{lstlisting}

%%%%%
\hide{
Self index::take_snapshot(){
	return Self(self);
}
}
%%%%%

% In this scenario, the graph index is organized as a tree-of-array.
% A PaC-Tree~\cite{dhulipala2022pac} in CPAM is used to store the vertices and leverage its multi-version capability.
% The edges are associated with the tree node of its vertex, stored in form of a \parray{}, introduced in \cref{sec:design:chrono}.\yan{fill in}

We use two structures to maintain the snapshots, \emph{chrono prefix array} and \emph{tree of prefix array}.
The latter also utilizes \parray{} to compress edges but employs a PAM tree~\cite{Sun2018PAM} to manage the vertices,
where each vertex is a tree node associated with a \parray{}.
Compared to \emph{chrono prefix array}, the PAM-based structure supports the snapshot branching and concurrent access
in a single-writer multiple-reader manner, bringing more functionality at a higher cost.
We compare their performance in \cref{sec:eval:snapshots}.

Implementing either structure requires a shim layer for the graph interface,
and any prebuilt algorithm such as \vamana{} can be adapted afterwards without modification.

\subsection{Range-Filtered Search}\label{app:range}
%The separation of \emph{algorithm} from \emph{data structure} can also be used to improve existing ANN systems. 
%\annlib{} can also be used to improve existing ANN systems. 
%We demonstrate this through a case study on \emph{range-filtered} ANNS, in which every point in $P$ carries a scalar attribute (e.g., a timestamp), 
%and every query has a range $[l,h]$. The task is to find the \knn{} of $q$ among the points with attributes in $[l,h]$.
%WinFilter~\cite{engels2024approximate} and UNIFY~\cite{liang2024unify} are two recent systems for this problem, developed independently and neither designed around \annlib{}.
In this section, we demonstrate how \annlib{} can be used to improve existing ANN systems through a case study on \emph{range-filtered} ANNS. 
In this task, every point in $P$ carries a scalar attribute (e.g., a timestamp), and every query has a range $[l,h]$. The task is to find the \knn{} of $q$ among the points with attributes in $[l,h]$.
We select two recent solutions, WinFilter~\cite{engels2024approximate} and UNIFY~\cite{liang2024unify}. 
%WinFilter sorts the points by attribute and recursively splits them into equally sized parts, forming a \emph{window search tree}. 
%Every node of that tree carries its own ANN index, built over the points the node covers.
WinFilter employs a \emph{window search tree}. Given points sorted by attribute, it builds a global ANN graph and then recursively partitions the points into equal halves, constructing an ANN graph for each subset. 
This hierarchy of ANN graphs allows any query range to be covered by a combination of $O(\log n)$ graphs. 
UNIFY partitions the points by attribute into $B$ equally sized segments and builds a single graph over them, in which every point carries one neighbor list per segment plus a global list marked out of those per-segment edges.
\hide{A query searches only the segments its range covers, falling back to the global list once the range spans nearly the whole dataset.}

While both methods prove effective for range-filter search, a closer examination of their implementations reveals that each carries a weakness orthogonal to its core contribution: 
WinFilter features fast parallel construction but a redundant graph container, resulting in high memory usage for storing multiple graphs; 
UNIFY employs a compact graph container but a less efficient lock-based construction algorithm. 
%Such pairings of a strong algorithm with a weak container, or the converse, recur throughout ANN research, and they confound comparison across systems, 
%since measuring one implementation against another reports the gap between two complete packages rather than the relative merit of the underlying designs. 
Such cases often confound comparison across ANN systems, since the gap comes from two complete packages rather than the relative merit of the underlying designs. 
Through clean modularization, \annlib{} also facilitates fair, controlled benchmarking. 
Specifically, we adapt \annlib{}'s \emph{parallel data structures} to WinFilter to resolve its memory overhead, and its \emph{parallel algorithms} to UNIFY to boost parallelism. 
As such, an optimization made available to one system is equally available to the other. 
In both cases, we achieved improved performance, and also obtained a fair comparison between them. 
%each system retains its own algorithmic design, while WinFilter adopts \annlib{}'s graph container and UNIFY adopts \annlib{}'s construction algorithm.
%the remainder of both systems also shares a common substrate, namely the same distance kernels, parallel scheduler, and allocator. 
%Each change below therefore affects only one of \annlib{}'s two layers, and an optimization made available to one system is equally available to the other.
%As such, each substitution can be evaluated without interference from unrelated implementation differences. 

\myparagraph{WinFilter.} WinFilter is built on ParlayANN~\cite{manohar2024parlayann}, an ANN library with efficient parallel construction. 
Its window search tree, however, materializes an independent index object at each tree node, every one of which carries its own map from point IDs to a local, node-relative numbering, 
so a point is stored logarithmic times along its root-to-leaf path.
The neighbor lists retained at adjacent levels frequently overlap, but the original implementation does not exploit this redundancy, which accounts for its memory overhead.
%Because \annlib{} decouples the graph container from the algorithm operating over it (\cref{sec:design:graph}), we replace each of these per-node maps with a single linearized store keyed by \texttt{(point, level)}, leaving the \vamana{} build and search code unchanged.
We replace each of these per-node maps with a single linearized store keyed by \texttt{(point, level)}, leaving the \vamana{} build and search code unchanged.
The store reuses two graph containers \annlib{} already provides: \texttt{nested\_array} (\cref{design:nested-array}) and \texttt{chrono\_prefix} (\cref{design:chrono-prefix-array}, \cref{fig:chrono}), 
with tree level substituted for version. 
Using version-aware data structures, our implementation substantially reduces memory usage while keeping comparable time efficiency (\cref{sec:eval:range}).
%Because only the container differs, the two traverse an identical search path and return bit-identical recall, so the cost of a more compact container can be measured directly rather than estimated (\cref{sec:eval:range}).
%The container is the \texttt{graph\_t} of \annlib{}'s descriptor, so substituting either of these, or a different container entirely, requires a one-line change (\cref{sec:design:overview}).

\myparagraph{UNIFY.} As noted above, UNIFY's global graph is a marked subset of the per-segment edges rather than a separate copy, so the baseline already exhibits a compact memory footprint.
Its build is parallel, but derives that parallelism from fine-grained locking: points are inserted concurrently, and each insertion acquires a per-vertex lock on every neighbor list it modifies.
Build throughput is therefore limited by contention on these locks.
%We instead reconstruct the same graph over \annlib{}'s existing single-layer \vamana{} (whose $\alpha$-pruning substitutes for the layer hierarchy) and its batch-parallel build: intra-segment graphs are constructed over complete segments, cross-segment links are subsequently added against those completed graphs, and the resulting structure is symmetrized, all without per-point locking.
We instead reconstruct the same graph over \annlibvamana{} and its batch-parallel build: intra-segment graphs are constructed over complete segments, cross-segment links are subsequently added against those completed graphs, and the resulting structure is symmetrized, all lock-free. 
This therefore reduces build time substantially (\cref{sec:eval:range}). 
%The result is not isomorphic to the original construction, but it replaces lock-mediated parallelism with batch parallelism and thereby reduces build time substantially (\cref{sec:eval:range}). 
%The choice of \vamana{} for the reconstruction is itself only one instance of a swappable algorithm module: the same batch-parallel build and segment logic would operate unchanged over \hnsw{} or any other algorithm in \annlib{}, again a one-line change to the same descriptor (\cref{sec:design:overview}).
% (whose $\alpha$-pruning substitutes for the layer hierarchy)

% \subsection{Streaming Query}\label{app:stm-query}
%\todo{give pseudo-code} 

\section{Experimental Evaluations}\label{sec:eval}
We evaluate \annlib{} and show that the flexibility for development comes without compromising performance. 
%Specifically, we conducted experiments for the batch insertion, regular vector search, batch deletion (presented in the supplemental material), filtered search, and snapshots.
Our results show that the performance of \annlib{} is generally comparable to or better than that of its monolithic counterparts.
\annlib{} is implemented as a C++ header-only library and uses ParlayLib~\cite{blelloch2020toolkit} as its default parallel framework to support work-stealing parallelism.

% We ran experiments on a machine equipped with four-way Intel\textregistered{} Xeon\textregistered{} Gold 6252 CPUs at 2.1GHz, and 1.5TiB main memory in 24 channels at 2933MT. 
We run experiments on a machine equipped with four-way
Intel\textregistered{} Xeon\textregistered{} Gold 6252 CPUs at $2.1$GHz, and
$1.5$TiB main memory in $24$ channels at $2933$MT/s.
The billion-scale experiments in \cref{sec:eval:rsearch} instead ran on a machine equipped with two-way Intel\textregistered{} Xeon\textregistered{}
Gold 6438Y+ CPUs at $2.0$GHz in two NUMA nodes, and $1$TiB main memory.
The datasets we used in our experiments are summarized in \cref{exp:dataset}.
\begin{figure*}[t]
  \centering\vspace{0.2em}
  \includegraphics[width=\linewidth]{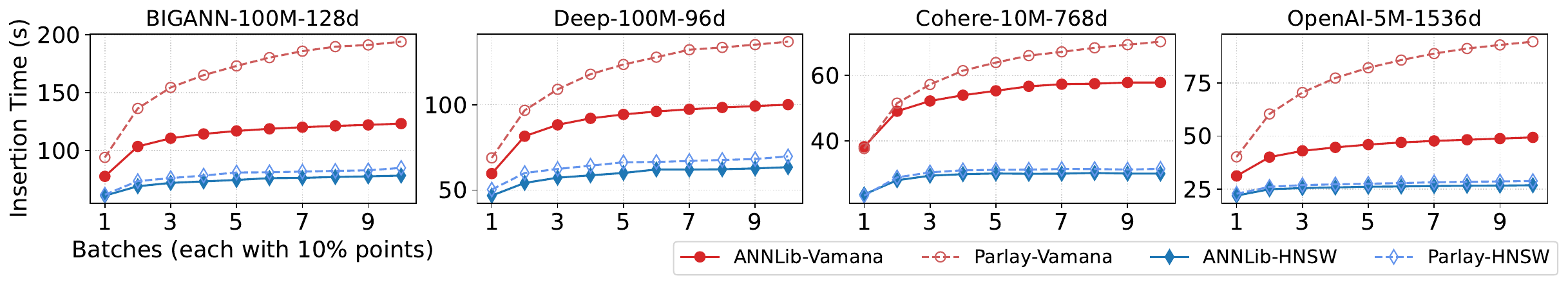}
  \caption{
\textbf{Incremental Insertion Time. Lower is better.}
x-axis indicates ten batches to insert; y-axis shows the per-batch insertion time.
  \label{fig:eval:ins}
  \vspace{-0.5em}
  }
\end{figure*}

\begin{table}[h]
\centering
\setlength{\tabcolsep}{4pt}
\renewcommand{\arraystretch}{1.05}
% \scriptsize
\footnotesize
\begin{tabular}{lccc|ccc}
\hline
\multirow{2}{*}{Name} & \multirow{2}{*}{Dim.} & \multirow{2}{*}{Size} & \multirow{2}{*}{Metrics} & \multicolumn{3}{c}{Labels} \\ \cline{5-7}
                      &                       &                       &                          & Type      & Num.      & Lbl/pt    \\ \hline
BIGANN~\cite{amsaleg2010datasets}          & 128   & 100M/1B  & L2   & Synthetic & 50        & 3.39    \\
DEEP~\cite{babenko2016efficient}           & 96    & 100M/1B  & L2   & ---  & ---  & ---\\
Cohere~\cite{cohere_wikipedia_embeddings}  & 768   & 10M   & cosine   & Synthetic & 50        & 3.39    \\
OpenAI~\cite{neelakantan2022text}           & 1,536  & 5M    & cosine   & ---  & ---  & --- \\
MARCO~\cite{chen2024ms}                    & 768   & 10M   & L2   & Native    & 10        & 6.13    \\
YFCC~\cite{thomee2016yfcc100m}              & 192   & 10M   & L2   & Native    & 200K   & 10.82   \\ \hline
\end{tabular}
\caption{Datasets used for experiments. 
%\emph{Synthetic} means the dataset is used for filtered search with labels generated in a Zipfian distribution ($\alpha=1$), \emph{Native} means that the dataset natively comes with metadata as filtering label can be directly used, and $\times$ means not being used for filtered search. 
For labels, \emph{Num.} is the number of distinct labels; \emph{Lbl/pt}: average \# of labels for each point.}
\label{exp:dataset}
\end{table}

\begin{figure*}[t]
  \centering
  \includegraphics[width=\linewidth]{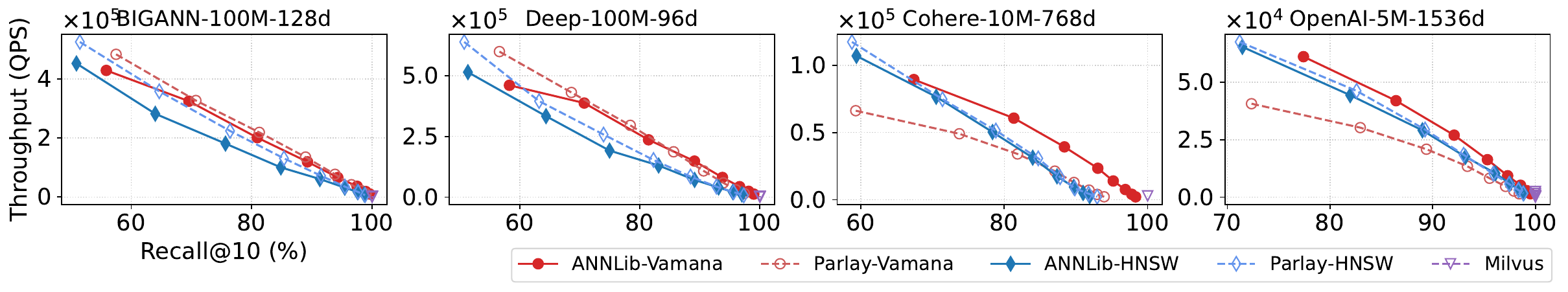}
  \caption{
  \textbf{QPS-recall of regular query. Top-right is better.}
  The optimal throughput-recall tradeoff over parameter sweeping.
  \label{fig:eval:rquery}
  \vspace{-0.5em}
}  
\end{figure*}

\hide{
\myparagraph{Measurement.}
We report batch operation time as the primary performance metric for our insertion, deletion, and snapshot tests. 
For snapshots, we measure the number of stored edges and total memory consumption. 
For search, we present recall-QPS curves as the primary results.
We use throughput rather than latency because throughput better reflects performance and scalability on a multi-core machine, while latency remains within an acceptable range. 
The points on each curve correspond to a Pareto optimum over the search parameter sweep after building the ANN indices.
}

\myparagraph{Baselines.}
We compare \annlib{} against \textit{Vamana} and \textit{HNSW} in ParlayANN~\cite{manohar2024parlayann} on construction and batch insertions. 
ParlayANN is known for fast construction and high query throughput, but does not support deletion or versioning. 
We compare batch deletion time with FreshDiskANN~\cite{singh2021fresh}. 
For filtered search, we compare to DiskANN~\cite{subramanya2019diskann}'s original implementation, 
and two industrial vector databases, Milvus~\cite{wang2021milvus} and Weaviate~\cite{weaviate2024}. 
For range-filter search, we compare with original implementations WinFilter and UNIFY. 
In all experiments, we focus only on the in-memory setting and exclude disk-related operations for both \annlib{} and baselines.

\subsection{Evaluating Insertion}
\label{sec:eval:ins}
We evaluate batch insertion for both \textit{Vamana} and \textit{HNSW} compared with \textit{ParlayANN}.
\hide{
Because index construction for these algorithms proceeds via incremental insertion, insertion time also reflects overall build performance. 
\Cref{fig:eval:ins} reports results on four datasets with varying sizes and dimensions.
For each dataset, we partition the points into ten equal-sized batches and, starting from an empty index, incrementally insert each batch. 
We plot the per-batch insertion time.
Overall, \annlib{} achieves shorter insertion times than \textit{ParlayANN}---which is known for fast builds---for both \textit{Vamana} and \textit{HNSW} across all datasets. 
}
% \edit{
Since these indices are built by incremental insertion,
insertion time also reflects build performance. 
We split each dataset into ten equal batches,
insert them into an initially empty index, and plot per-batch insertion time in \cref{fig:eval:ins}.
On all datasets, \annlib{} achieves shorter total time than \textit{ParlayANN} (itself known for fast builds) for both \textit{Vamana} and \textit{HNSW}.
% }
Our \textit{Vamana} is $1.4\times$ faster on average, up to $1.9\times$ on the last batch of \textit{OpenAI}.
The \textit{HNSW} on \annlib{} keeps runtime comparable to \parlayhnsw{}, with a $1.06\times$ speedup on average.
This improvement primarily stems from copy elision: our interfaces adhere to the true data requirements and avoid unnecessary copies.

% Each subfigure in \cref{sec:eval:ins} presents the time of continuously performing 10 batch insertions, each containing 10\% points from the dataset.
% insert the dataset by 10 batches, each containing 10\% points, and show the time of every single insertion.
% Each subfigure in \cref{sec:eval:ins} presents the per-batch time of incremental insertion with ten equally-divided batches.
% The insertion time grows with the index size increases.
% We observe that, compared to ParlayANN, our implementations plotted in solid lines achieve better performance of the \textit{Vamana} insertion on BIGANN, Cohere, and OpenAI datasets, and maintain similar performance in the rest of the cases.
% Specifically, our implementations takes xx per-batch insertion time on \textit{Vamana}, and xx time on \textit{HNSW}.
% Since the graph structures and algorithms used in both libraries are similar, the insertion performance is expected to be close, apart from negligible abstract overhead.
% We account for the ParlayANN's performance deterioration on ultra-high-dimensional datasets as the results of costly distance computation.

\subsection{Evaluating Scalability}
\label{sec:eval:scale}
\begin{figure*}[t]
\centering
\includegraphics[width=\linewidth]{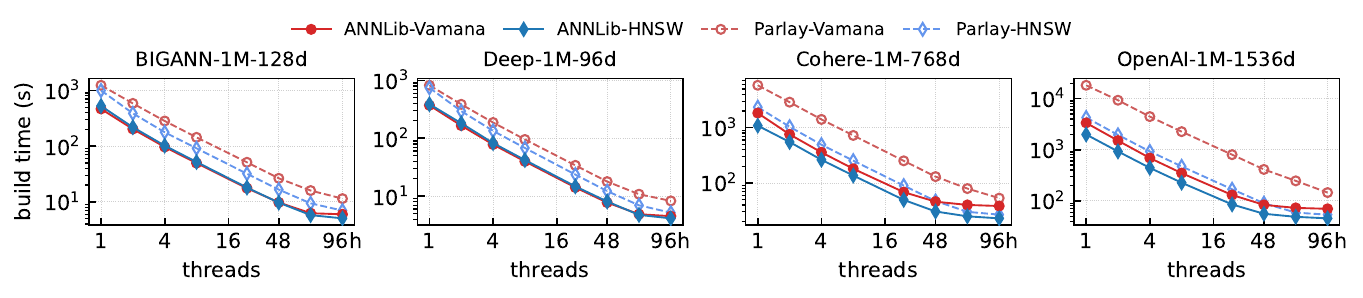}
\caption{\textbf{Build scalability with thread count. Lower is better.} Total time to build an index over 1M points, on log-log axes. Solid lines are \annlib{}, dashed lines are \textit{ParlayANN}. 96h: 96 cores with hyperthreading.\label{fig:scalability}}
\end{figure*}

% Data: experiments/results/scalability/ours+parlayann.tsv
We further assess how construction cost varies with the degree of parallelism available.
\Cref{fig:scalability} reports the time to directly build an index over 1M points using different thread counts.
Here \vamana{} uses $m=16$ instead of $32$ as in the other experiments, to reduce the experiment time.
Both libraries scale consistently across all four datasets and for both \vamana{} and \hnsw{}, and their build times remain comparable throughout the range. 
The rate of improvement diminishes beyond roughly 48 threads in both cases, as is expected once the work available per thread becomes small relative to synchronization and memory bandwidth.

\subsection{Evaluating Regular Vector Search}
\label{sec:eval:rsearch}
We evaluate both graph quality and search efficiency by running standard vector search over the built \textit{Vamana} and \textit{HNSW} indices.
\Cref{fig:eval:rquery} presents QPS-recall trade-offs for 10-NN search.
On \textit{Cohere} and \textit{OpenAI}, \annlibvamana{} is consistently faster than \parlayvamana{}, while their \hnsw{} curves are similar.
On \textit{BIGANN} and \textit{DEEP}, \annlib{} shows lower throughput in low-recall regions but converges to comparable QPS as recall approaches 1.0.
The low-recall throughput gap is due to a specialized optimizer in \textit{ParlayANN} that caps the number of node visits when the search parameter is small.
We also include Milvus in these tests, which is a vector database optimized for high query quality.
Across parameter sweeps, Milvus consistently achieves near-perfect recall, but at the cost of significantly lower throughput.

\begin{figure}[t]
\centering
\includegraphics[width=\linewidth]{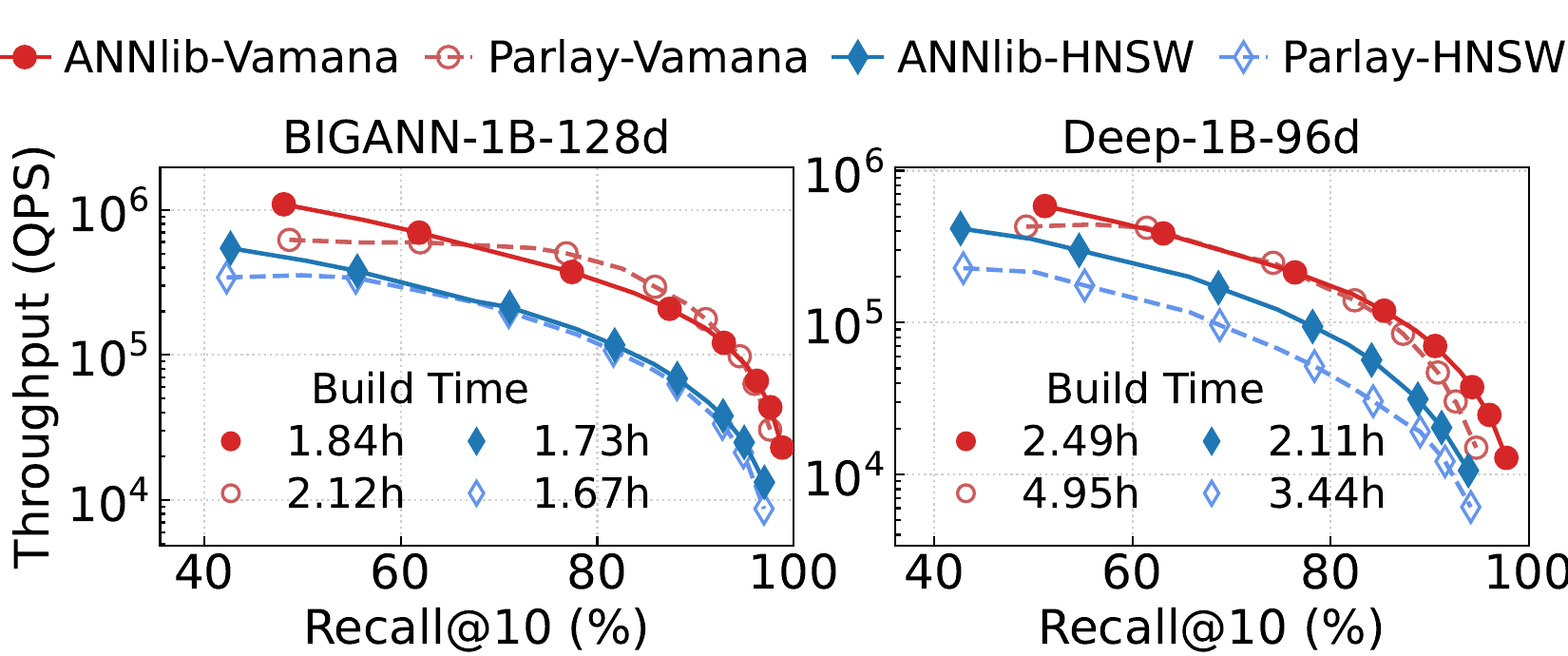}
\caption{\textbf{Build time (hours), QPS and recall for all algorithms on
billion-size datasets.}}
\label{fig:billion-qps-recall}
\end{figure}

\Cref{fig:billion-qps-recall} reports billion-scale results.
All indices build in a few hours on a single machine.
On DEEP-1B \annlib{} builds $1.99\times$ faster for \vamana{} and $1.63\times$ faster for \hnsw{}.
On BIGANN-1B it is $1.16\times$ faster for \vamana{}.
For search, \annlib{} leads throughout on DEEP-1B, where \annlibhnsw{} achieves $1.6$--$1.9\times$ the throughput of \parlayhnsw{}, and \annlibvamana{} reaches $1.4\times$ at $90.5\%$ recall and over $2\times$ beyond $94\%$.
On BIGANN-1B \annlibhnsw{} keeps $1.0$--$1.6\times$, while the \vamana{} curves cross.
ParlayANN's visited-node cap yields $1.35\times$ in $74\%$--$93\%$ recall, since at billion scale every avoided visit is a likely cache miss, but it also flattens the curve at small beams and caps recall earlier, so \annlib{} leads above $95\%$ and reaches $98.8\%$ against $97.6\%$.
% None of this required scalability-specific code: these are the same \vamana{} and \hnsw{} programs used throughout \cref{sec:eval}, just need to change the data path.

\hide{
We conduct a regular vector search over the built \textit{Vamana} indices to evaluate both graph quality and search efficiency. 
\Cref{fig:eval:rquery} presents the QPS-recall trade-offs for a 10-NN search.
Compared to \textit{ParlayANN}, our implementation shows similar curve on \textit{BIGANN}.
For \textit{DEEP} and \textit{OpenAI}, our implementation provides lower throughput in low-recall regions but approaches the same QPS when the query recall approaches to 1.0.
The throughput gap at low recall is caused by a special optimizer applied to \textit{ParlayANN} that limits the number of visits when the search parameter is small.
We also include a vector database, Milvus, in this tests, which targets providing high query quality in production scenarios.
Swapping over varying parameters, Milvus consistently achieves near perfect recall, however, at the cost of low throughput. }
%\zheqi{explanation}
% TODO: try add a zoom-in of the high-recall region to the figures

\subsection{Evaluating Deletion}
We use the algorithm in FreshDiskANN~\cite{singh2021fresh}, and compare our implementation to their code and other baselines.
We present the full performance evaluation in Appendix~\ref{app:bat-del-detail} and summarize the main conclusions here: our batch deletion algorithm built upon \annlib{}'s general interface achieves similar performance (both in deletion time and consequent query quality) to the original highly-optimized algorithm in FreshDiskANN, and are much better than other baselines. 
%We compare to FreshDiskANN~\cite{singh2021fresh} other baselines.
% In summary, our work built upon \annlib{}'s general interface achieves similar performance (both in deletion time and consequent query quality) to the highly-optimized deletion algorithm in FreshDiskANN~\cite{singh2021fresh}, and are much better than other baselines.

\begin{figure}[ht]
  \centering
  \includegraphics[width=\linewidth]{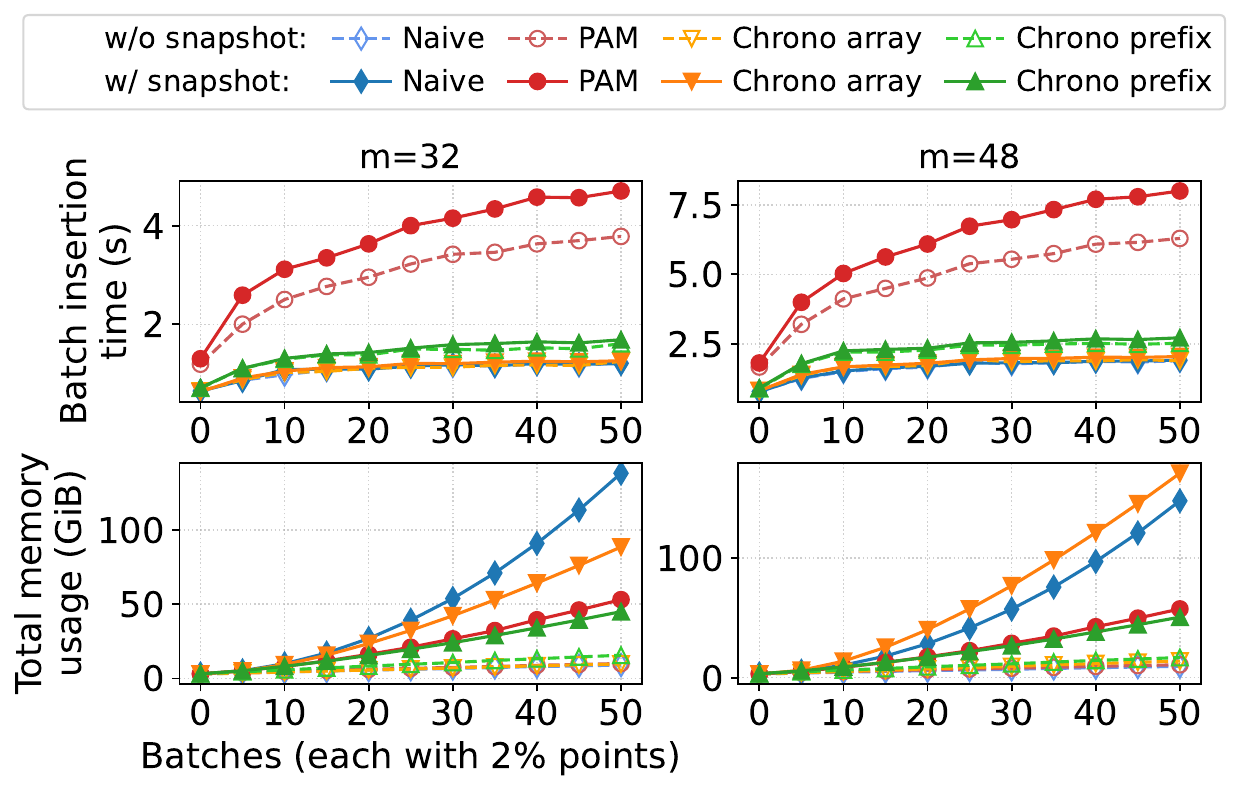}
%  \caption{
%  \textbf{Memory usage for snapshots. Lower is better. }
%  Incrementally insert 50 batches and monitor the memory usage in varying degree and base size.
%  \label{fig:snap-mem}}
  \caption{
  \textbf{Build time and memory usage for inserting 50 batches (2\% points each) on BIGANN-10M with/without snapshots. Lower is better.}
  %Per-batch build time of incrementally inserting 50 batches and saving all historical versions. 
  ``w/o snapshot'': only keep the last version. ``w/ snapshot'': keep all historical versions. 
  ``Naive'': baseline solutions, which updates an array in-place if no snapshot is taken, otherwise performs a deep copy for each snapshot.
  \label{fig:snap-buildtime}}
\end{figure}

\subsection{Evaluating Filtered Search}
\label{sec:eval:fsearch}
% \begin{figure*}[h]
%   \centering
%   \includegraphics[width=\linewidth]{figures/filtered_search.bigann-10M}
%   \caption{QPS-recall of filtered search on BIGANN-10M}
% \end{figure*}
% \begin{figure*}[h]
%   \centering
%   \includegraphics[width=\linewidth]{figures/filtered_search.deep-10M}
%   \caption{QPS-recall of filtered search on DEEP-10M}
% \end{figure*}
% \begin{figure}[h]
%   \centering
%   \includegraphics[width=.9\linewidth]{figures/filtered_search.marco-10M}
%   \caption{QPS-recall of filtered search on MARCO-10M}
% \end{figure}
% \begin{figure*}[h]
%   \centering
%   \includegraphics[width=\linewidth]{figures/filtered_search.yfcc-10M}
%   \caption{QPS-recall of filtered search on YFCC-10M}
% \end{figure*}
%Filter ANNS represents another milestone for \annlib{}, delivering satisfactory performance. 

We present the QPS--recall curves in Appendix~\ref{app:filter}, and summarize the conclusion here.
We follow the settings of Filtered-DiskANN and evaluate filtered search on a multi-label base set with a single filter per query, over four datasets under varying specificity.
In summary, \annlib{} matches or outperforms the evaluated monolithic solutions in most cases, while also simplifying system development.

\hide{
The filtering strategy during index construction involves identifying pairs of label sets with non-empty intersections (i.e., $L_u\cap L_v\neq \emptyset$); at query time, the algorithm directly checks whether a label set of index contains the singleton query label (i.e., $l\in L_u$).
We implement metadata filtering using the \textit{like} operation provided by Milvus and Weaviate vector databases by concatenating each base point's label set into a comma-separated string. 
Therefore, label strings are considered as the metadata associated with each point. 
Milvus utilities metadata filtering before ANN search by restricting the search scope to entities that meet the specified criteria, thereby enhancing search performance. Weaviate supports hybrid search that integrates both keyword and vector search; users can adjust the hybrid search balance by tuning the parameter $\alpha$ from 0 to 1. 
As no keywords are present in our experiments, we set $\alpha$ to 1 to achieve a pure vector search complemented by metadata filtering.}

\subsection{Evaluating Snapshots}
\label{sec:eval:snapshots}
The techniques in \cref{sec:design:graph} avoid duplicating vertices and edges across snapshots, thereby reducing memory usage.
To evaluate this, we incrementally construct an index preserving all historical versions and report both total build time and memory usage in \cref{fig:snap-buildtime}.
\hide{
As the number of snapshots grows, memory consumption increases; however, our prefix-sharing structure significantly reduces usage---by up to \(70.4\%\)---relative to a chronological list scheme, across diverse datasets and parameter settings.
The benefit is especially pronounced for datasets with larger degree bounds, where avoiding full edge copies yields greater savings.
Also, the combination of a parallel tree with a prefix array substantially reduces memory, at the cost of longer build time.
As shown in \cref{fig:snap-buildtime}, enabling the tree structure increases build time by \(1.13\times\) to \(3.32\times\) compared to the naïve scheme.
Enabling snapshot operations adds a further \(1.22\times\) overhead.
}
% \edit{
% As the number of snapshots grows, memory consumption increases.
Across diverse datasets and settings,
our prefix-sharing structure cuts memory usage by up to \(70.4\%\) over a chronological-list scheme.
The benefit is especially pronounced for datasets with larger degree bounds, where avoiding full edge copies matters more.
As shown in \cref{fig:snap-buildtime}, pairing a parallel tree with the prefix array saves further memory at the cost of build time:
enabling the tree structure increases build time by \(1.13\times\) to \(3.33\times\) over the naive scheme,
and enabling snapshots adds another \(1.22\times\).
% }
Due to \annlib{}'s modular design, developers can easily swap graph components to target their specific application scenarios and achieves various trade-offs.

\subsection{Evaluating Range-Filtered Search}
\label{sec:eval:range}
\begin{figure*}[!t]
\centering
\footnotesize
% Plain tabular rather than tabular*: with \extracolsep{\fill} the stretch glue
% lands on one side of a vertical rule only, leaving it off-centre in the gap.
\setlength{\tabcolsep}{8pt}
\renewcommand{\arraystretch}{1.15}
% booktabs pads above/below its rules, which would break the vertical rules
\setlength{\aboverulesep}{0pt}
\setlength{\belowrulesep}{0pt}
\begin{tabular}{
  l|
  *{5}{l}|
  *{5}{l}
}
\toprule
\multirow{3}{*}{Dataset}
& \multicolumn{5}{c|}{Build time (s)}
& \multicolumn{5}{c}{Peak resident memory (GiB)} \\
\cmidrule(lr){2-6} \cmidrule(lr){7-11}
& \multicolumn{3}{c}{WinFilter} & \multicolumn{2}{c|}{UNIFY}
& \multicolumn{3}{c}{WinFilter} & \multicolumn{2}{c}{UNIFY} \\
\cmidrule(lr){2-4} \cmidrule(lr){5-6} \cmidrule(lr){7-9} \cmidrule(lr){10-11}
& Orig. & \texttt{nested} & \texttt{chrono} & Orig. & ours
& Orig. & \texttt{nested} & \texttt{chrono} & Orig. & ours \\
\midrule
BIGANN & 19.08  & 18.03  & 18.02  & 195.09  & 27.95  & 3.08  & 2.46 & 2.12 & 2.59  & 1.88  \\
DEEP   & 23.48  & 23.73  & 23.32  & 141.52  & 35.64  & 3.92  & 2.80 & 2.49 & 2.36  & 2.30  \\
Cohere & 201.15 & 166.15 & 161.71 & 434.45  & 283.53 & 11.20 & 7.71 & 7.34 & 7.36  & 7.11  \\
OpenAI & 498.54 & 386.75 & 388.86 & 1377.39 & 513.66 & 20.01 & 13.44 & 13.14 & 13.07 & 13.06 \\
\midrule
\textbf{Geomean}
       & 81.87  & 72.41  & 71.70  & \textbf{358.52}  & \textbf{109.75} & \textbf{7.21}  & \textbf{5.17} & \textbf{4.75} & 4.92  & 4.48  \\
\bottomrule
\end{tabular}

\vspace{0.3em}
{\footnotesize \textbf{(a) Build cost. Lower is better.}\par}
\vspace{0.2em}

\includegraphics[width=\linewidth]{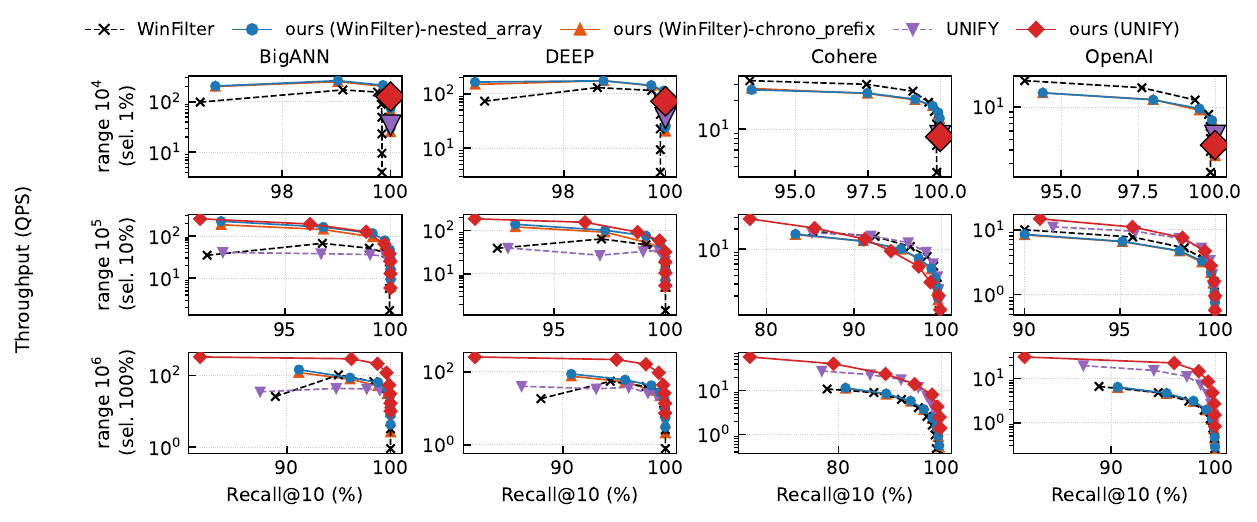}

\vspace{0.1em}
{\footnotesize \textbf{(b) Query throughput vs.\ recall. Top-right is better.}\par}
\vspace{0.6em}

\caption{Range-filtered ANN: WinFilter and UNIFY against their \annlib{} re-implementations (\cref{app:range}). In (a), \texttt{nested} and \texttt{chrono} are our \texttt{nested\_array} and \texttt{chrono\_prefix} edge-storage containers for WinFilter, and \texttt{ours} under UNIFY re-implements its graph on \annlibvamana{}'s batch-parallel builder. In (b), one row per query-range width and one column per dataset, dashed lines are the published baselines and solid lines their re-implementations. At range $10^4$ both UNIFY variants dispatch to an exact pre-filter scan regardless of \texttt{ef}, so each is a single black-outlined marker at (recall $100\%$, median QPS) rather than a curve.\label{fig:range-cost-qps}}
\end{figure*}

We re-implement WinFilter~\cite{engels2024approximate} and UNIFY~\cite{liang2024unify} inside \annlib{} as described in \cref{app:range}, and compare each against its own published code at matched parameters, on four one-million-point datasets with a synthetic scalar attribute drawn uniformly at random. \Cref{fig:range-cost-qps}(a) reports build time and peak resident memory, and \cref{fig:range-cost-qps}(b) the query throughput.

%For WinFilter, build time improves slightly on the baseline (roughly $12\%$ in the geometric mean), while peak memory falls by $1.25$--$1.57\times$: the single linearized store of \cref{app:range} replaces thousands of per-node point-ID maps. In the geometric mean, \texttt{chrono\_\allowbreak prefix} uses roughly $8\%$ less memory than \texttt{nested\_\allowbreak array}. Because substituting the graph container leaves the search algorithm unchanged, the QPS-recall curves (\cref{fig:range-cost-qps}(b)) closely track WinFilter's own reported throughput.
As expected, the peak memory of our implementation falls by $1.25$--$1.57\times$ compared to its original version: the single linearized store of \cref{app:range} replaces thousands of per-node point-ID maps. 
On average, \texttt{chrono\_\allowbreak prefix} uses roughly $8\%$ less memory than \texttt{nested\_\allowbreak array}. 
Crucially, this memory reduction comes without sacrificing performance: our QPS-recall curves (\cref{fig:range-cost-qps}(b)) closely track WinFilter's original throughput, 
with the construction time even slightly better (roughly $12\%$ in the geometric mean). 

For UNIFY, build time improves on every dataset ($1.5$--$7.0\times$, geometric mean $3.27\times$), consistent with the replacement of its lock-based concurrent \hnsw{} construction by \annlibvamana{}'s batch-parallel build, which removes per-vertex locking. 
We present the corresponding scalability study in Appendix~\ref{app:range-scale}.
It shows that the re-implementation is already $1.2$--$3.1\times$ faster on a single thread, where neither side has parallelism to exploit, and that removing the locks lets it keep scaling to the full machine while UNIFY itself can stall once hyperthreads are engaged.

\Cref{fig:range-cost-qps}(b) plots query throughput against recall at matched beam width (ef from 10 to 1280), at range widths drawn uniformly at random from $[1, 10^4]$, $[1, 10^5]$, and $[1, 10^6]$ out of $10^6$ total points. At range $10^4$ (selectivity up to $1\%$), UNIFY and its \annlib{} re-implementation both dispatch to their exact pre-filter scan (\cref{app:range}), so recall is always exactly $100\%$ there and \texttt{ef} does not affect the search path. At that range both consequently reduce to a single operating point rather than a sweep, and are plotted as one marker.

Our QPS-recall curve is broadly comparable to each baseline, with no systematic regression from substituting either the graph container or the construction algorithm. Because these two case studies target build time and memory specifically, we did not otherwise tune \annlib{}'s search-time parameters beyond matching the baselines' own settings.

\hide{The techniques proposed in \cref{sec:design:graph} avoid storing duplicated vertices and edges among snapshots and save the memory. 
To evaluate this, we test a setting that construct an index incrementally, and preserve all history versions. 
We report both the total time and the memory usage in \cref{fig:snap-mem,fig:snap-buildtime}.
Memory consumption increases with the number of snapshots increases. 
Our implementation, with the prefix-sharing structure, significantly reduces memory consumption up to 70.4\% compared to the regular chronological list scheme, among varying datasets and the parameter settings. 
Especially for datasets with a larger degree bound, the benefits from avoiding full edge copies is more pronounced.

The combination of the parallel tree and the prefix array greatly saves memory, at the cost of longer build time. 
Shown in \cref{fig:snap-buildtime}, enabling the tree structure increases the build time from $1.13\times$ to $3.32\times$, compared to the naive scheme. Applying the snapshot operations brings another $1.22\times$ increment of the build time. With \annlib{}'s highly modular design, the developer can swap the graph components depending on their target scenario to obtain the best tradeoff.

} 

\section{Related Work}
\label{sec:related-work}

% ANNS has been widely studied. For page limit, we present a full literature review in the supplemental material, and only review the most relevant ones here. 
% ANNS has been studied extensively. Due to the page limit, we present a full literature review in the supplemental material and review only the most relevant work here.
%For page limit, we review the most relevant previous work here. A full literature review is presented in the supplemental material.

\myparagraph{ANNS Algorithms.}
A broad range of studies has addressed ANNS algorithms, including DiskANN~\cite{subramanya2019diskann}, Stars~\cite{carey2022star}, SPFresh~\cite{xu2023spfresh}, ParlayANN~\cite{manohar2024parlayann}, and other graph-based approaches~\cite{malkov2020hnsw,fu2019nsg,munoz2019hcnng,fu2021high,mcinnes2020pynndescent,zhang2022hierarchical,ren2020hmann,luo2025aversearch,gottesbueren2025unleashing,rubel2026pipnn,zhang2025cleann}. These state-of-the-art systems handle billion-scale datasets on a single node, but their codebases have grown to over 10,000 lines of code~\cite{subramanya2019diskann}, making it hard to locate and modify relevant parts to port features and tailor functions. Recently, kANNolo~\cite{delfino2025kannolo} introduced a Rust library with composable abstractions for ANNS with varying vector density and quantization methods.

\myparagraph{Vector Databases.} ANNS is widely used in databases for vector queries, known as vector databases. Vector databases either extend traditional engines~\cite{pgvector,zhang2023vbase} or build specialized systems~\cite{wang2021milvus}. They typically wrap a few fixed ANNS algorithms, requiring substantial code modifications to integrate peripheral database capabilities.

\myparagraph{Filtered and Range-Filtered ANNS.}
%There are a host of ANNS related applications, including filtered search, dynamic update, and out-of-distribution query.
Filtered ANNS is an extension of ANN search by retrieving the top-$k$ similar results while satisfying a specified set of metadata predicates.
% There are several existing work on filtered ANNS. We categorize recent filtered ANNS approaches in 2 categories by the ingredients of ANN indexes.
Existing work on filtered ANNS can be broadly classified into two main categories. 
The first category treats the filtering operation as an independent step, without modifying the original ANNS indexes. 
Filtering can be applied either before (pre-filtering)~\cite{patel2024acorn,weaviate2024,wang2021milvus,wei2020analyticdb,chen2024singlestore} or after (post-filtering)~\cite{yang2020pase,douze2024faiss, johnson2019billion} the vector search.
However, both pre- and post-filtering can be inefficient, especially under highly selective filtering conditions.
%Pre-filtering first predicts on filtering conditions by building an independent index for each potential filters, then perform the vector search while evaluating the filters on candidates. Systems like ACORN~\cite{patel2024acorn}, Weaviate~\cite{weaviate2024}, Milvus~\cite{wang2021milvus}, AnalyticDB-V~\cite{wei2020analyticdb}, SingleStore-V~\cite{chen2024singlestore}, and MSVBASE~\cite{zhang2023vbase} adopt this approach. 
%However, because it is hard to predict the filtering patterns and elements that will be contained in the search, this approach scales poorly, becoming inefficient for higher filtering labels on large datasets. 
%By contrast, post-filtering first runs a standard ANN query to retrieve a larger candidate set than the target $k$. Then, it scans this set to identify those that match the specified attributes. This %technique, implemented in systems like PASE~\cite{yang2020pase} and FAISS~\cite{douze2024faiss, johnson2019billion}, is straightforward to deploy. However, it becomes highly inefficient when the filter is very selective because the vast majority of retrieved candidates are discarded, resulting in significant wasted computation.
The second builds filter-aware indexes that integrate filtering information into the ANNS index. This approach modifies the index construction and search algorithms to simultaneously handle vector similarity and metadata constraints, as in Filtered-DiskANN~\cite{gollapudi2023filtered}, or combines attribute constraints with the index, such as Native Hybrid Query (NHQ)~\cite{wang2023efficient} and MA-NSW~\cite{xu2020multiattribute}. %There also exist IVF-based approach for filtered search.
%IVF$^2$ introduces a label-centric IVF-based (inverted file indexing) approach to apply the Vamana index and an IVF index according to the query specificity.
Range-filtered ANNS replaces the label predicate with a numeric range over a scalar attribute~\cite{zuo2024serf,engels2024approximate,xu2025irangegraph,liang2024unify,wang2026wow,liu2026rrann}, where the central design question is how to amortize a single index over the many ranges a query may ask for.
\Cref{app:range} re-implements two of these systems, WinFilter and UNIFY, on \annlib{}.

\section{Conclusions}
We propose \textbf{\annlib{}}, a library for graph-based ANNS development. 
The goal of \annlib{} is to achieve both \emp{high performance} and \emp{flexible functionality}. 
We carefully decouple and independently optimize both the \emph{algorithm} and the \emph{data structure} components in an ANNS system. 
Both of them allow for easy integration of state-of-the-art solutions as modules, as well as our new designs. 
We show how these components can be composed to address sophisticated settings, such as filtered search, historical queries on snapshots, and range-filter search. 
\hide{Our experiment show that solutions built atop \annlib{}'s general interfaces achieve comparable or mostly even better performance as previous work specifically for each application. }
Our experiments show that solutions built atop \annlib{}'s general interfaces match or outperform previous work designed specifically for each application.
%We release our anonymous code at ~\cite{annlib-pub}. 

\clearpage

%-------------------------------------------------------------------------------
\bibliographystyle{ACM-Reference-Format}
\bibliography{ann,annlib}

@string{neurips = "Annual Conference on Neural Information Processing Systems (NeurIPS)"}

@string{icml = "Machine Learning, Proceedings of the Twenty-Third International Conference (ICML)"}

@string{sisap = "Similarity Search and Applications (SISAP)"}

@inproceedings{peng2023iqan,
  title={iQAN: Fast and Accurate Vector Search with Efficient Intra-Query Parallelism on Multi-Core Architectures},
  author={Peng, Zhen and Zhang, Minjia and Li, Kai and Jin, Ruoming and Ren, Bin},
  booktitle={Proceedings of the 28th ACM SIGPLAN Annual Symposium on Principles and Practice of Parallel Programming},
  pages={313--328},
  year={2023}
}

@article{peng2022speed,
  title={Speed-ANN: Low-Latency and High-Accuracy Nearest Neighbor Search via Intra-Query Parallelism},
  author={Peng, Zhen and Zhang, Minjia and Li, Kai and Jin, Ruoming and Ren, Bin},
  journal={arXiv preprint arXiv:2201.13007},
  year={2022}
}

@inproceedings{raoofy2023overcoming,
  title={Overcoming Weak Scaling Challenges in Tree-Based Nearest Neighbor Time Series Mining},
  author={Raoofy, Amir and Karlstetter, Roman and Schreiber, Martin and Trinitis, Carsten and Schulz, Martin},
  booktitle={International Conference on High Performance Computing},
  pages={317--338},
  year={2023},
  organization={Springer}
}

@inproceedings{annbench,
	title={ANN-Benchmarks},
	author={ANN Benchmarks Authors},
  year={2023},
	url={https://ann-benchmarks.com/index.html}
}

@inproceedings{mou2017refined,
  author       = {Wenlong Mou and
                  Liwei Wang},
  title        = {A Refined Analysis of {LSH} for Well-dispersed Data Points},
  booktitle    = {Proceedings of the Fourteenth Workshop on Analytic Algorithmics and
                  Combinatorics, {ANALCO} 2017, Barcelona, Spain, Hotel Porta Fira,
                  January 16-17, 2017},
  pages        = {174--182},
  publisher    = {{SIAM}},
  year         = {2017},
  url          = {https://doi.org/10.1137/1.9781611974775.18},
  doi          = {10.1137/1.9781611974775.18}
}

@inproceedings{subramanya2019diskann,
  author    = {Suhas Jayaram Subramanya and
               Fnu Devvrit and
               Harsha Vardhan Simhadri and
               Ravishankar Krishnaswamy and
               Rohan Kadekodi},
  title     = {DiskANN: Fast Accurate Billion-point Nearest Neighbor Search on a
               Single Node},
  booktitle = neurips,
  pages     = {13748--13758},
  year      = {2019}
}

@article{malkov2020hnsw,
  author    = {Yury A. Malkov and
               Dmitry A. Yashunin},
  title     = {Efficient and Robust Approximate Nearest Neighbor Search Using Hierarchical
               Navigable Small World Graphs},
  journal   = {{IEEE} Trans. Pattern Anal. Mach. Intell.},
  volume    = {42},
  number    = {4},
  pages     = {824--836},
  year      = {2020}
}

@article{fu2019nsg,
  author    = {Cong Fu and
               Chao Xiang and
               Changxu Wang and
               Deng Cai},
  title     = {Fast Approximate Nearest Neighbor Search With The Navigating Spreading-out
               Graph},
  journal   = {Proc. {VLDB} Endow.},
  volume    = {12},
  number    = {5},
  pages     = {461--474},
  year      = {2019}
}

@article{munoz2019hcnng,
  author    = {Javier Alvaro Vargas Mu{\~{n}}oz and
               Marcos Andr{\'{e}} Gon{\c{c}}alves and
               Zanoni Dias and
               Ricardo da Silva Torres},
  title     = {Hierarchical Clustering-Based Graphs for Large Scale Approximate Nearest
               Neighbor Search},
  journal   = {Pattern Recognit.},
  volume    = {96},
  year      = {2019}
}

@article{fu2021high,
  author    = {Cong Fu and
               Changxu Wang and
               Deng Cai},
  title     = {High dimensional similarity search with satellite system graph:
               Efficiency, scalability, and unindexed query compatibility},
  journal   = {{IEEE} Trans. Pattern Anal. Mach. Intell.},
  volume    = {44},
  number    = {8},
  pages     = {4139--4150},
  year      = {2022}
}

@article{zhang2022hierarchical,
  author    = {Jiaru Zhang and
               Ruhui Ma and
               Tao Song and
               Yang Hua and
               Zhengui Xue and
               Chenyang Guan and
               Haibing Guan},
  title     = {Hierarchical Satellite System Graph for Approximate Nearest Neighbor
               Search on Big Data},
  journal   = {ACM/IMS Trans. Data Sci.},
  volume    = {2},
  number    = {4},
  year      = {2022}
}

@string{spaa = "{ACM} Symposium on Parallelism in Algorithms
	and Architectures (SPAA)"}

@article{lu2021hvs,
  author    = {Kejing Lu and
               Mineichi Kudo and
               Chuan Xiao and
               Yoshiharu Ishikawa},
  title     = {HVS: Hierarchical Graph Structure Based on Voronoi Diagrams for Solving
               Approximate Nearest Neighbor Search},
  journal   = {Proc. VLDB Endow.},
  volume    = {15},
  number    = {2},
  pages     = {246–258},
  year      = {2022}
}

@inproceedings{harwood2016fanng,
  author    = {Ben Harwood and
               Tom Drummond},
  title     = {Fanng: Fast approximate nearest neighbour graphs},
  booktitle = {Proceedings of the IEEE Conference on Computer Vision and Pattern
               Recognition},
  pages     = {5713--5722},
  year      = {2016}
}

@article{wang2021comprehensive,
  author    = {Mengzhao Wang and
               Xiaoliang Xu and
               Qiang Yue and
               Yuxiang Wang},
  title     = {A Comprehensive Survey and Experimental Comparison of Graph-Based
               Approximate Nearest Neighbor Search},
  journal   = {Proc. {VLDB} Endow.},
  volume    = {14},
  number    = {11},
  pages     = {1964--1978},
  year      = {2021}
}

@misc{weaviate2024,
  author = {Etienne Dilocker and Bob van Luijt and Byron Voorbach and Mohd Shukri Hasan and Abdel Rodriguez and Dirk Alexander Kulawiak and Marcin Antas and Parker Duckworth},
  year = {2024},
  title = {Weaviate},
  howpublished  = "Webpage",
  url = {https://github.com/weaviate/weaviate}
}

@article{amuller2020ann,
  author    = {Martin Aum{\"{u}}ller and
               Erik Bernhardsson and
               Alexander John Faithfull},
  title     = {ANN-Benchmarks: {A} benchmarking tool for approximate nearest neighbor
               algorithms},
  journal   = {Information Systems},
  volume    = {87},
  year      = {2020}
}

@inproceedings{dong2011efficient,
  author    = {Wei Dong and
               Moses Charikar and
               Kai Li},
  editor    = {Sadagopan Srinivasan and
               Krithi Ramamritham and
               Arun Kumar and
               M. P. Ravindra and
               Elisa Bertino and
               Ravi Kumar},
  title     = {Efficient k-nearest neighbor graph construction for generic similarity
               measures},
  booktitle = {Proceedings of the 20th International Conference on World Wide Web (WWW)},
  pages     = {577--586},
  publisher = {{ACM}},
  year      = {2011}
}

@misc{mcinnes2020pynndescent,
  author        = "Leland McInnes",
  year          = "2020",
  title         = "PyNNDescent for Fast Approximate Nearest Neighbors",
  howpublished  = "Webpage",
  url           = "https://pynndescent.readthedocs.io/en/latest/",
  lastaccessed  = "December 15, 2022",
}

@inproceedings{beygelzimer2006cover,
  author    = {Alina Beygelzimer and
               Sham M. Kakade and
               John Langford},
  title     = {Cover trees for nearest neighbor},
  booktitle = icml,
  series    = {{ACM} International Conference Proceeding Series},
  volume    = {148},
  pages     = {97--104},
  publisher = {{ACM}},
  year      = {2006}
}

@inproceedings{gu2022parallel,
  title={Parallel Cover Trees and their Applications},
  author={Gu, Yan and Napier, Zachary and Sun, Yihan and Wang, Letong},
  booktitle=spaa,
  pages={259--272},
  year={2022}
}

@article{pham2022falconn,
  author    = {Ninh Pham and
               Tao Liu},
  title     = {Falconn++: {A} Locality-sensitive Filtering Approach for Approximate
               Nearest Neighbor Search},
  journal   = {CoRR},
  volume    = {abs/2206.01382},
  year      = {2022},
 doi       = {10.48550/arXiv.2206.01382},
  eprinttype = {arXiv},
  eprint    = {2206.01382}
}

@article{singh2021fresh,
  author    = {Aditi Singh and
               Suhas Jayaram Subramanya and
               Ravishankar Krishnaswamy and
               Harsha Vardhan Simhadri},
  title     = {FreshDiskANN: {A} Fast and Accurate Graph-Based {ANN} Index for Streaming
               Similarity Search},
  journal   = {CoRR},
  volume    = {abs/2105.09613},
  year      = {2021},
  url       = {https://arxiv.org/abs/2105.09613},
  eprinttype = {arXiv},
  eprint    = {2105.09613}
}

@article{jaiswal2022ood,
  author    = {Shikhar Jaiswal and
               Ravishankar Krishnaswamy and
               Ankit Garg and
               Harsha Vardhan Simhadri and
               Sheshansh Agrawal},
  title     = {OOD-DiskANN: Efficient and Scalable Graph {ANNS} for Out-of-Distribution
               Queries},
  journal   = {CoRR},
  volume    = {abs/2211.12850},
  year      = {2022},
  url       = {https://doi.org/10.48550/arXiv.2211.12850},
  doi       = {10.48550/arXiv.2211.12850},
  eprinttype = {arXiv}
}

@inproceedings{iwasaki2016pruned,
  author    = {Masajiro Iwasaki},
  title     = {Pruned Bi-directed K-nearest Neighbor Graph for Proximity Search},
  booktitle = sisap,
  series    = {Lecture Notes in Computer Science},
  volume    = {9939},
  pages     = {20--33},
  year      = {2016}
}

@inproceedings{chen2021spann,
  author    = {Qi Chen and
               Bing Zhao and
               Haidong Wang and
               Mingqin Li and
               Chuanjie Liu and
               Zengzhong Li and
               Mao Yang and
               Jingdong Wang},
  title     = {{SPANN:} Highly-efficient Billion-scale Approximate Nearest Neighborhood
               Search},
  booktitle = neurips,
  pages     = {5199--5212},
  year      = {2021}
}

@manual{chen2018sptag,
  author    = {Qi Chen and
               Haidong Wang and
               Mingqin Li and
               Gang Ren and
               Scarlett Li and
               Jeffery Zhu and
               Jason Li and
               Chuanjie Liu and
               Lintao Zhang and
               Jingdong Wang},
  title     = {SPTAG: A library for fast approximate nearest neighbor search},
  url       = {https://github.com/Microsoft/SPTAG},
  year      = {2018}
}

@inproceedings{ren2020hmann,
  author    = {Jie Ren and
               Minjia Zhang and
               Dong Li},
  editor    = {Hugo Larochelle and
               Marc'Aurelio Ranzato and
               Raia Hadsell and
               Maria{-}Florina Balcan and
               Hsuan{-}Tien Lin},
  title     = {{HM-ANN:} Efficient Billion-Point Nearest Neighbor Search on Heterogeneous
               Memory},
  booktitle = neurips,
  year      = {2020}
}

@article{toussaint1980relative,
  author    = {Godfried T. Toussaint},
  title     = {The relative neighbourhood graph of a finite planar set},
  journal   = {Pattern recognition},
  volume    = {12},
  number    = {4},
  pages     = {261--268},
  year      = {1980}
}

@misc{hnswlib2019hnswlib,
  year          = "2019",
  title         = "Hnswlib - fast approximate nearest neighbor search",
  howpublished  = "Webpage",
  url           = "https://github.com/nmslib/hnswlib",
  lastaccessed  = "December 16, 2019",
}

@article{yesantharao2021parallel,
  author    = {Rahul Yesantharao and
               Yiqiu Wang and
               Laxman Dhulipala and
               Julian Shun},
  title     = {Parallel Batch-Dynamic kd-Trees},
  journal   = {CoRR},
  volume    = {abs/2112.06188},
  year      = {2021},
  url       = {https://arxiv.org/abs/2112.06188},
  eprinttype = {arXiv},
  eprint    = {2112.06188},
}

@inproceedings{dobson2022parallel,
  author    = {Guy E. Blelloch and
               Magdalen Dobson},
  title     = {Parallel Nearest Neighbors in Low Dimensions with Batch Updates},
  booktitle = {Proceedings of the Symposium on Algorithm Engineering and Experiments (ALENEX)},
  pages     = {195--208},
  publisher = {{SIAM}},
  year      = {2022},
  url       = {https://doi.org/10.1137/1.9781611977042.16},
  doi       = {10.1137/1.9781611977042.16},
}

@inproceedings{blelloch2020toolkit,
  author    = {Guy E. Blelloch and
               Daniel Anderson and
               Laxman Dhulipala},
  title     = {ParlayLib - {A} Toolkit for Parallel Algorithms on Shared-Memory Multicore
               Machines},
  booktitle = spaa,
  pages     = {507--509},
  publisher = {{ACM}},
  year      = {2020},
  url       = {https://doi.org/10.1145/3350755.3400254},
  doi       = {10.1145/3350755.3400254},
}

@article{xu2022proximity,
  author    = {Zhaozhuo Xu and
               Weijie Zhao and
               Shulong Tan and
               Zhixin Zhou and
               Ping Li},
  title     = {Proximity Graph Maintenance for Fast Online Nearest Neighbor Search},
  journal   = {CoRR},
  volume    = {abs/2206.10839},
  year      = {2022},
  url       = {https://doi.org/10.48550/arXiv.2206.10839},
  doi       = {10.48550/arXiv.2206.10839},
  eprinttype = {arXiv},
  eprint    = {2206.10839}
}

@misc{spacev2021spacev,
  author  = "SpaceV Contributors",
  year          = "2021",
  title         = "SPACEV1B: A billion-Scale vector dataset for text descriptors",
  howpublished  = "Webpage",
  url           = "https://github.com/microsoft/SPTAG/tree/main/datasets/SPACEV1B",
  lastaccessed  = "March 16, 2023",
}

@misc{baranchuk2021benchmarks,
  author  = "Dmitry Baranchuk and Artem Babenko",
  year          = "2021",
  title         = "Benchmarks for Billion-Scale Similarity Search",
  howpublished  = "Webpage",
  url           = "https://research.yandex.com/blog/benchmarks-for-billion-scale-similarity-search",
  lastaccessed  = "March 16, 2023",
}

@inproceedings{baranchuk2018revisiting,
  author    = {Dmitry Baranchuk and
               Artem Babenko and
               Yury Malkov},
  title     = {Revisiting the Inverted Indices for Billion-Scale Approximate Nearest
               Neighbors},
  booktitle = {Computer Vision - {ECCV} 2018},
  series    = {Lecture Notes in Computer Science},
  volume    = {11216},
  pages     = {209--224},
  publisher = {Springer},
  year      = {2018}
}

@misc{chase2023vector,
	title={Vector DB text generation},
	url={https://python.langchain.com/en/latest/modules/chains/index_examples/vector_db_text_generation.html},
	journal={Vector DB Text Generation - LangChain 0.0.128},
	author={Chase, Harrison},
	year={2023},
	month={Mar}
}

@misc{tawalke2023pr,
	title={PR: SK Vectordb Connector Work - Merging forked branch; PR from Fork to SK Branch by Tawalke · pull request 83 · Microsoft/Semantic-Kernel},
	url={https://github.com/microsoft/semantic-kernel/pull/83},
	journal={GitHub},
	author={tawalke},
	year={2023},
	month={Mar}
}

@misc{bing,
  title={Microsoft Bing Search Engine},
	url={https://www.bing.com/new},
	journal={Bing},
	publisher={Microsoft},
  year={2023},
}

@misc{stallbaumer2023copilot,
	title={Introducing Microsoft 365 copilot},
	url={https://www.microsoft.com/en-us/microsoft-365/blog/2023/03/16/introducing-microsoft-365-copilot-a-whole-new-way-to-work/},
	 journal={Microsoft 365 Blog},
	 author={Stallbaumer, Colette},
	year={2023},
	month={Mar}
}

@misc{pinecone, url={https://www.pinecone.io/}, title={Pinecone: Vector Database for Vector Search}, year={2023}}

@misc{weaviate, url={https://weaviate.io/}, title={Weaviate: The AI Native Vector Database}, year={2023}}

@misc{lucene,
  title = {Apache Lucene},
  url={https://lucene.apache.org/},
  year = {2023}}

@inproceedings{dhulipala2022pac,
author = {Dhulipala, Laxman and Blelloch, Guy E. and Gu, Yan and Sun, Yihan},
title = {PaC-trees: supporting parallel and compressed purely-functional collections},
year = {2022},
isbn = {9781450392655},
publisher = {Association for Computing Machinery},
address = {New York, NY, USA},
doi = {10.1145/3519939.3523733},
pages = {108–121},
location = {San Diego, CA, USA},
series = {PLDI 2022}
}

@inproceedings{overwijk2022clueweb22,
  title={ClueWeb22: 10 billion web documents with rich information},
  author={Overwijk, Arnold and Xiong, Chenyan and Callan, Jamie},
  booktitle={Proceedings of the 45th international ACM SIGIR conference on research and development in information retrieval},
  pages={3360--3362},
  year={2022}
}

@misc{eirinaki2018recommender,
  title={Recommender systems for large-scale social networks: A review of challenges and solutions},
  author={Eirinaki, Magdalini and Gao, Jerry and Varlamis, Iraklis and Tserpes, Konstantinos},
  journal={Future generation computer systems},
  volume={78},
  pages={413--418},
  year={2018},
  publisher={Elsevier}
}

@article{lecun1995convolutional,
  title={Convolutional networks for images, speech, and time series},
  author={LeCun, Yann and Bengio, Yoshua and others},
  journal={The handbook of brain theory and neural networks},
  volume={3361},
  number={10},
  pages={1995},
  year={1995},
  publisher={Citeseer}
}

@article{mikolov2013distributed,
  title={Distributed representations of words and phrases and their compositionality},
  author={Mikolov, Tomas and Sutskever, Ilya and Chen, Kai and Corrado, Greg S and Dean, Jeff},
  journal={Advances in neural information processing systems},
  volume={26},
  year={2013}
}

@article{scarselli2008graph,
  title={The graph neural network model},
  author={Scarselli, Franco and Gori, Marco and Tsoi, Ah Chung and Hagenbuchner, Markus and Monfardini, Gabriele},
  journal={IEEE transactions on neural networks},
  volume={20},
  number={1},
  pages={61--80},
  year={2008},
  publisher={IEEE}
}

@inproceedings{tagami2017annexml,
  title={Annexml: Approximate nearest neighbor search for extreme multi-label classification},
  author={Tagami, Yukihiro},
  booktitle={Proceedings of the 23rd ACM SIGKDD international conference on knowledge discovery and data mining},
  pages={455--464},
  year={2017}
}

@inproceedings{zhang2022uni,
  title={Uni-retriever: Towards learning the unified embedding based retriever in bing sponsored search},
  author={Zhang, Jianjin and Liu, Zheng and Han, Weihao and Xiao, Shitao and Zheng, Ruicheng and Shao, Yingxia and Sun, Hao and Zhu, Hanqing and Srinivasan, Premkumar and Deng, Weiwei and others},
  booktitle={Proceedings of the 28th ACM SIGKDD Conference on Knowledge Discovery and Data Mining},
  pages={4493--4501},
  year={2022}
}

@inproceedings{zhang2018visual,
  title={Visual search at alibaba},
  author={Zhang, Yanhao and Pan, Pan and Zheng, Yun and Zhao, Kang and Zhang, Yingya and Ren, Xiaofeng and Jin, Rong},
  booktitle={Proceedings of the 24th ACM SIGKDD international conference on knowledge discovery \& data mining},
  pages={993--1001},
  year={2018}
}

@inproceedings{huang2020embedding,
  title={Embedding-based retrieval in facebook search},
  author={Huang, Jui-Ting and Sharma, Ashish and Sun, Shuying and Xia, Li and Zhang, David and Pronin, Philip and Padmanabhan, Janani and Ottaviano, Giuseppe and Yang, Linjun},
  booktitle={Proceedings of the 26th ACM SIGKDD International Conference on Knowledge Discovery \& Data Mining},
  pages={2553--2561},
  year={2020}
}

@article{patel2024acorn,
  title={Acorn: Performant and predicate-agnostic search over vector embeddings and structured data},
  author={Patel, Liana and Kraft, Peter and Guestrin, Carlos and Zaharia, Matei},
  journal={Proceedings of the ACM on Management of Data},
  volume={2},
  number={3},
  pages={1--27},
  year={2024},
  publisher={ACM New York, NY, USA}
}

@article{chen2024singlestore,
  title={SingleStore-V: An Integrated Vector Database System in SingleStore},
  author={Chen, Cheng and Jin, Chenzhe and Zhang, Yunan and Podolsky, Sasha and Wu, Chun and Wang, Szu-Po and Hanson, Eric and Sun, Zhou and Walzer, Robert and Wang, Jianguo},
  journal={Proceedings of the VLDB Endowment},
  volume={17},
  number={12},
  pages={3772--3785},
  year={2024},
  publisher={VLDB Endowment}
}

@article{wang2023efficient,
  title={An efficient and robust framework for approximate nearest neighbor search with attribute constraint},
  author={Wang, Mengzhao and Lv, Lingwei and Xu, Xiaoliang and Wang, Yuxiang and Yue, Qiang and Ni, Jiongkang},
  journal={Advances in Neural Information Processing Systems},
  volume={36},
  pages={15738--15751},
  year={2023}
}

@article{xu2020multiattribute,
  title={Multiattribute approximate nearest neighbor search based on navigable small world graph},
  author={Xu, Xiaoliang and Li, Chang and Wang, Yuxiang and Xia, Yixing},
  journal={Concurrency and Computation: Practice and Experience},
  volume={32},
  number={24},
  pages={e5970},
  year={2020},
  publisher={Wiley Online Library}
}

@article{jafari2021survey,
  title={A survey on locality sensitive hashing algorithms and their applications},
  author={Jafari, Omid and Maurya, Preeti and Nagarkar, Parth and Islam, Khandker Mushfiqul and Crushev, Chidambaram},
  journal={arXiv preprint arXiv:2102.08942},
  year={2021}
}

@inproceedings{datar2004locality,
  title={Locality-sensitive hashing scheme based on p-stable distributions},
  author={Datar, Mayur and Immorlica, Nicole and Indyk, Piotr and Mirrokni, Vahab S},
  booktitle={Proceedings of the twentieth annual symposium on Computational geometry},
  pages={253--262},
  year={2004}
}

@inproceedings{gionis1999similarity,
  title={Similarity search in high dimensions via hashing},
  author={Gionis, Aristides and Indyk, Piotr and Motwani, Rajeev and others},
  booktitle={Vldb},
  volume={99},
  number={6},
  pages={518--529},
  year={1999}
}

@article{babenko2014inverted,
  title={The inverted multi-index},
  author={Babenko, Artem and Lempitsky, Victor},
  journal={IEEE transactions on pattern analysis and machine intelligence},
  volume={37},
  number={6},
  pages={1247--1260},
  year={2014},
  publisher={IEEE}
}

@article{wang2025accelerating,
  title={Accelerating Graph Indexing for ANNS on Modern CPUs},
  author={Wang, Mengzhao and Wu, Haotian and Ke, Xiangyu and Gao, Yunjun and Zhu, Yifan and Zhou, Wenchao},
  journal={arXiv preprint arXiv:2502.18113},
  year={2025}
}

@inproceedings{radford2021learning,
  title={Learning transferable visual models from natural language supervision},
  author={Radford, Alec and Kim, Jong Wook and Hallacy, Chris and Ramesh, Aditya and Goh, Gabriel and Agarwal, Sandhini and Sastry, Girish and Askell, Amanda and Mishkin, Pamela and Clark, Jack and others},
  booktitle={International conference on machine learning},
  pages={8748--8763},
  year={2021},
  organization={PmLR}
}

@inproceedings{dhulipala2019ctree,
author = {Dhulipala, Laxman and Blelloch, Guy E. and Shun, Julian},
title = {Low-latency graph streaming using compressed purely-functional trees},
year = {2019},
publisher = {Association for Computing Machinery},
doi = {10.1145/3314221.3314598},
pages = {918--934},
series = {PLDI 2019}
}

@inproceedings{Sun2018PAM,
title = { PAM: Parallel Augmented Maps },
author = { Sun, Yihan and  Ferizovic, Daniel and  Blelloch, Guy E.},
booktitle = {ACM Symposium on Principles and Practice of Parallel Programming (PPoPP)},
year = { 2018 }
}

@misc{zhang2025cleann,
  author = {Ziyu Zhang and Yuanhao Wei and Joshua Engels and Julian Shun},
  title  = {{CleANN}: Efficient Full Dynamism in Graph-based Approximate Nearest Neighbor Search},
  howpublished = {arXiv:2507.19802},
  year   = {2025}
}

@misc{luo2025aversearch,
  author = {Jingjia Luo and Mingxing Zhang and Kang Chen and Xia Liao and Yingdi Shan and Jinlei Jiang and Yongwei Wu},
  title  = {Efficient Graph-Based Approximate Nearest Neighbor Search Achieving: Low Latency Without Throughput Loss},
  howpublished = {arXiv:2504.20461},
  year   = {2025}
}

@article{gottesbueren2025unleashing,
  author  = {Lars Gottesb{\"u}ren and Laxman Dhulipala and Rajesh Jayaram and Jakub {\L}{\k{a}}cki},
  title   = {Unleashing Graph Partitioning for Large-Scale Nearest Neighbor Search},
  journal = {Proc. {VLDB} Endow.},
  volume  = {18},
  number  = {6},
  pages   = {1649--1662},
  year    = {2025}
}

@misc{rubel2026pipnn,
  author = {Tobias Rubel and Richard Wen and Laxman Dhulipala and Lars Gottesb{\"u}ren and Rajesh Jayaram and Jakub {\L}{\k{a}}cki},
  title  = {{PiPNN}: Ultra-Scalable Graph-Based Nearest Neighbor Indexing},
  howpublished = {arXiv:2602.21247},
  year   = {2026}
}

@article{men2025pkdtree,
  author  = {Ziyang Men and Zheqi Shen and Yan Gu and Yihan Sun},
  title   = {Parallel kd-tree with Batch Updates},
  journal = {Proc. {ACM} Manag. Data},
  volume  = {3},
  number  = {1},
  year    = {2025}
}

@inproceedings{ding2026skiplist,
  author    = {Xiangyun Ding and Rohin Garg and Yan Gu and Yihan Sun},
  title     = {Parallel Metric Skip Lists and Nearest Neighbor Search},
  booktitle = {{ACM} Symposium on Parallelism in Algorithms and Architectures (SPAA)},
  year      = {2026}
}

@inproceedings{yu2025pecann,
  author    = {Shangdi Yu and Joshua Engels and Yihao Huang and Julian Shun},
  title     = {{PECANN}: Parallel Efficient Clustering with Graph-Based Approximate Nearest Neighbor Search},
  booktitle = {{SIAM} Conference on Applied and Computational Discrete Algorithms (ACDA)},
  year      = {2025}
}

@inproceedings{dhulipala2024muvera,
  author    = {Laxman Dhulipala and Majid Hadian and Rajesh Jayaram and Jason Lee and Vahab Mirrokni},
  title     = {{MUVERA}: Multi-Vector Retrieval via Fixed Dimensional Encodings},
  booktitle = {Advances in Neural Information Processing Systems (NeurIPS)},
  year      = {2024}
}

@inproceedings{wang2021milvus,
	title={Milvus: A purpose-built vector data management system},
	author={Wang, Jianguo and Yi, Xiaomeng and Guo, Rentong and Jin, Hai and Xu, Peng and Li, Shengjun and Wang, Xiangyu and Guo, Xiangzhou and Li, Chengming and Xu, Xiaohai and others},
	booktitle={Proceedings of the 2021 International Conference on Management of Data},
	pages={2614--2627},
	year={2021}
}

@article{guo2022manu,
	title={Manu: A Cloud Native Vector Database Management System},
	author={Guo, Rentong and Luan, Xiaofan and Xiang, Long and Yan, Xiao and Yi, Xiaomeng and Luo, Jigao and Cheng, Qianya and Xu, Weizhi and Luo, Jiarui and Liu, Frank and others},
	journal={VLDB Endowment},
	pages={3548},
	year={2022}
}

@inproceedings{gollapudi2023filtered,
	author = {Gollapudi, Siddharth and Karia, Neel and Sivashankar, Varun and Krishnaswamy, Ravishankar and Begwani, Nikit and Raz, Swapnil and Lin, Yiyong and Zhang, Yin and Mahapatro, Neelam and Srinivasan, Premkumar and Singh, Amit and Simhadri, Harsha Vardhan},
	title = {Filtered-DiskANN: Graph Algorithms for Approximate Nearest Neighbor Search with Filters},
	year = {2023},
	booktitle = {Proceedings of the ACM Web Conference 2023},
	pages = {3406–3416},
	numpages = {11},
	series = {WWW '23}
}

@inproceedings{manohar2024parlayann,
	author = {Manohar, Magdalen Dobson and Shen, Zheqi and Blelloch, Guy and Dhulipala, Laxman and Gu, Yan and Simhadri, Harsha Vardhan and Sun, Yihan},
	title = {ParlayANN: Scalable and Deterministic Parallel Graph-Based Approximate Nearest Neighbor Search Algorithms},
	year = {2024},
	booktitle = {Proceedings of the 29th ACM SIGPLAN Annual Symposium on Principles and Practice of Parallel Programming},
	pages = {270–285},
	numpages = {16},
	series = {PPoPP '24}
}

@inproceedings{zhang2023vbase,
	author = {Qianxi Zhang and Shuotao Xu and Qi Chen and Guoxin Sui and Jiadong Xie and Zhizhen Cai and Yaoqi Chen and Yinxuan He and Yuqing Yang and Fan Yang and Mao Yang and Lidong Zhou},
	title = {{VBASE}: Unifying Online Vector Similarity Search and Relational Queries via Relaxed Monotonicity},
	booktitle = {17th USENIX Symposium on Operating Systems Design and Implementation (OSDI 23)},
	year = {2023},
	pages = {377--395},
	publisher = {USENIX Association}
}

@inproceedings{yang2020pase,
	author = {Yang, Wen and Li, Tao and Fang, Gai and Wei, Hong},
	title = {PASE: PostgreSQL Ultra-High-Dimensional Approximate Nearest Neighbor Search Extension},
	year = {2020},
	pages = {2241–2253},
	numpages = {13},
	series = {SIGMOD '20}
}

@article{johnson2019billion,
	title={Billion-scale similarity search with GPUs},
	author={Johnson, Jeff and Douze, Matthijs and J{\'e}gou, Herv{\'e}},
	journal={IEEE Transactions on Big Data},
	volume={7},
	number={3},
	pages={535--547},
	year={2019},
	publisher={IEEE}
}

@article{douze2024faiss,
	title={The faiss library},
	author={Douze, Matthijs and Guzhva, Alexandr and Deng, Chengqi and Johnson, Jeff and Szilvasy, Gergely and Mazar{\'e}, Pierre-Emmanuel and Lomeli, Maria and Hosseini, Lucas and J{\'e}gou, Herv{\'e}},
	journal={arXiv preprint arXiv:2401.08281},
	year={2024}
}

@inproceedings{xu2023spfresh,
	title={SPFresh: Incremental In-Place Update for Billion-Scale Vector Search},
	author={Xu, Yuming and Liang, Hengyu and Li, Jin and Xu, Shuotao and Chen, Qi and Zhang, Qianxi and Li, Cheng and Yang, Ziyue and Yang, Fan and Yang, Yuqing and others},
	booktitle={Proceedings of the 29th Symposium on Operating Systems Principles},
	pages={545--561},
	year={2023}
}

@article{wei2020analyticdb,
	title={AnalyticDB-V: a hybrid analytical engine towards query fusion for structured and unstructured data},
	author={Wei, Chuangxian and Wu, Bin and Wang, Sheng and Lou, Renjie and Zhan, Chaoqun and Li, Feifei and Cai, Yuanzhe},
	journal={Proceedings of the VLDB Endowment},
	volume={13},
	number={12},
	pages={3152--3165},
	year={2020},
	publisher={VLDB Endowment}
}

@inproceedings{carey2022star,
	author = {Carey, CJ and Halcrow, Jonathan and Jayaram, Rajesh and Mirrokni, Vahab and Schudy, Warren and Zhong, Peilin},
	booktitle = {Advances in Neural Information Processing Systems},
	pages = {21470--21481},
	title = {Stars: Tera-Scale Graph Building for Clustering and Learning},
	volume = {35},
	year = {2022}
}

@misc{pgvector,
	author        = {Andrew Kane},
	year          = "2022",
	title         = "pgvector",
	howpublished  = "Webpage",
	url           = "https://github.com/pgvector/pgvector/",
	lastaccessed  = "July, 2023"
}

@misc{amsaleg2010datasets,
  author       = {Laurent Amsaleg and Hervé Jegou},
  title        = {Datasets for approximate nearest neighbor search},
  year         = {2010},
  howpublished = {\url{http://corpus-texmex.irisa.fr/}},
  note         = {[Online; accessed 20-May-2018]}
}

@inproceedings{babenko2016efficient,
  title={Efficient indexing of billion-scale datasets of deep descriptors},
  author={Babenko, Artem and Lempitsky, Victor},
  booktitle={Proceedings of the IEEE Conference on Computer Vision and Pattern Recognition},
  pages={2055--2063},
  year={2016}
}

@misc{cohere_wikipedia_embeddings,
  author       = {Cohere},
  title        = {Wikipedia (en) embedded with cohere.ai multilingual-22-12 encoder},
  year         = {2023},
  howpublished = {\url{https://huggingface.co/datasets/Cohere/wikipedia-22-12-en-embeddings}},
  note         = {[Online; accessed 22-January-2025]}
}

@article{neelakantan2022text,
  title={Text and code embeddings by contrastive pre-training},
  author={Neelakantan, Arvind and Xu, Tao and Puri, Raul and Radford, Alec and Han, Jesse Michael and Tworek, Jerry and Yuan, Qiming and Tezak, Nikolas and Kim, Jong Wook and Hallacy, Chris and others},
  journal={arXiv preprint arXiv:2201.10005},
  year={2022}
}

@inproceedings{chen2024ms,
  title={MS MARCO Web Search: a Large-scale Information-rich Web Dataset with Millions of Real Click Labels},
  author={Chen, Qi and Geng, Xiubo and Rosset, Corby and Buractaon, Carolyn and Lu, Jingwen and Shen, Tao and Zhou, Kun and Xiong, Chenyan and Gong, Yeyun and Bennett, Paul and others},
  booktitle={Companion Proceedings of the ACM on Web Conference 2024},
  pages={292--301},
  year={2024}
}

@article{thomee2016yfcc100m,
  title={Yfcc100m: The new data in multimedia research},
  author={Thomee, Bart and Shamma, David A and Friedland, Gerald and Elizalde, Benjamin and Ni, Karl and Poland, Douglas and Borth, Damian and Li, Li-Jia},
  journal={Communications of the ACM},
  volume={59},
  number={2},
  pages={64--73},
  year={2016},
  publisher={ACM New York, NY, USA}
}

@article{simhadri2024results,
  title={Results of the Big ANN: NeurIPS'23 competition},
  author={Simhadri, Harsha Vardhan and Aum{\"u}ller, Martin and Ingber, Amir and Douze, Matthijs and Williams, George and Manohar, Magdalen Dobson and Baranchuk, Dmitry and Liberty, Edo and Liu, Frank and Landrum, Ben and others},
  journal={arXiv preprint arXiv:2409.17424},
  year={2024}
}

@inproceedings{li2018JD,
  author = {Li, Jie and Liu, Haifeng and Gui, Chuanghua and Chen, Jianyu and Ni, Zhenyuan and Wang, Ning and Chen, Yuan},
  title = {The Design and Implementation of a Real Time Visual Search System on JD E-commerce Platform},
  year = {2018},
  publisher = {Association for Computing Machinery},
  doi = {10.1145/3284028.3284030},
  booktitle = {Proceedings of the 19th International Middleware Conference Industry},
  pages = {9–16}
}

@article{xi2025vecflow,
  title={VecFlow: A High-Performance Vector Data Management System for Filtered-Search on GPUs},
  author={Xi, Jingyi and Mo, Chenghao and Karsin, Benjamin and Chirkin, Artem and Li, Mingqin and Zhang, Minjia},
  journal={arXiv preprint arXiv:2506.00812},
  year={2025}
}

@article{lin2025survey,
  title={Survey of Filtered Approximate Nearest Neighbor Search over the Vector-Scalar Hybrid Data},
  author={Lin, Yanjun and Zhang, Kai and He, Zhenying and Jing, Yinan and Wang, X Sean},
  journal={arXiv preprint arXiv:2505.06501},
  year={2025}
}

@article{zuo2024serf,
  title={SeRF: segment graph for range-filtering approximate nearest neighbor search},
  author={Zuo, Chaoji and Qiao, Miao and Zhou, Wenchao and Li, Feifei and Deng, Dong},
  journal={Proceedings of the ACM on Management of Data},
  volume={2},
  number={1},
  pages={1--26},
  year={2024},
  publisher={ACM New York, NY, USA}
}

@article{engels2024approximate,
  title={Approximate nearest neighbor search with window filters},
  author={Engels, Joshua and Landrum, Benjamin and Yu, Shangdi and Dhulipala, Laxman and Shun, Julian},
  journal={arXiv preprint arXiv:2402.00943},
  year={2024}
}

@article{liang2024unify,
  title={UNIFY: Unified Index for Range Filtered Approximate Nearest Neighbors Search},
  author={Liang, Anqi and Zhang, Pengcheng and Yao, Bin and Chen, Zhongpu and Song, Yitong and Cheng, Guangxu},
  journal={Proceedings of the VLDB Endowment},
  volume={18},
  number={4},
  pages={1118--1130},
  year={2024}
}

@misc{annlib-pub,
   title = {{ANNlib: A Development Framework for Efficient Approximate Nearest Neighbor Search}},
   howpublished = {\url{https://github.com/ucrparlay/ANNlib-pub}},
   version = {0.1},
   year = {2025}
}

@article{xu2025irangegraph,
  author={Xu, Yuexuan and Gao, Jianyang and Gou, Yutong and Long, Cheng and Jensen, Christian S.},
  title={{iRangeGraph}: Improvising Range-dedicated Graphs for Range-filtering Nearest Neighbor Search},
  journal={Proceedings of the ACM on Management of Data},
  volume={2},
  number={6},
  year={2024}
}

@article{wang2026wow,
  author={Wang, Ziqi and Zhang, Jingzhe and Hu, Wei},
  title={{WoW}: A Window-to-Window Incremental Index for Range-Filtering Approximate Nearest Neighbor Search},
  journal={Proceedings of the ACM on Management of Data},
  year={2026}
}

@article{liu2026rrann,
  author={Liu, Yingfan and Wu, Tong and Xie, Jiadong and Zhao, Yang and Yu, Jeffrey Xu and Cui, Jiangtao},
  title={Generalized Range Filtering Approximate Nearest Neighbor Search: Containment and Overlap},
  journal={arXiv preprint arXiv:2605.26474},
  year={2026}
}

@inproceedings{delfino2025kannolo,
	author = {Delfino, Leonardo and Erriquez, Domenico and Martinico, Silvio and Nardini, Franco Maria and Rulli, Cosimo and Venturini, Rossano},
	title = {kANNolo: Sweet and Smooth Approximate k-Nearest Neighbors Search},
	year = {2025},
	publisher = "Springer Nature Switzerland",
	isbn = {978-3-031-88716-1},
	doi = {10.1007/978-3-031-88717-8_29},
	pages = {400--406}
}

%-------------------------------------------------------------------------------
% Supplemental material (appendix + full related work), included in this version.
%-------------------------------------------------------------------------------
\clearpage
\appendix
\section{Filtered Search}\label{app:flt-search-detail}
We provide filtered ANNS as a built-in block in \annlib{} based on two algorithms in Filtered-DiskANN~\cite{gollapudi2023filtered}: \svamana{} and \fvamana{}.
\svamana{} consists of two steps that: 1) build a regular \vamana{} graph over points with each label; and 2) merge all the sub-graphs using \fprune{}.
By inheriting from \vamana{}, \svamana{} easily reuses the insertion implementation on the first step.
For the second step, we add a new function \texttt{g.merge(h)} for a \svamana{} graph \texttt{g} that merges (add all vertices and edges) from a graph \texttt{h}.
For vertices exist both in \texttt{g} and \texttt{h}, the edges from two sources are merged by \fprune{}, which is implemented by a custom \textit{f\_nbhs}, as shown in \cref{code:svamana}.
We also improve the performance by parallelizing the per-label graph merge process, where the original algorithm is constrained by the bottleneck of sequential merging.

\begin{lstlisting}[
	language=C++,
	caption={Stitched-Vamana Merge},
	morekeywords={foreach},
	label={code:svamana},
	escapechar=@
]
auto get_f_nbhs(seq labels){
	return [&](nid_t u, nid_t w){
		auto es = get_edges(u);
		auto check = filter([&](auto e){
			auto (u,v) = e;
			return u.label @$\cap$@ labels @$\neq \emptyset$@; 	});
		return es | check;  }; }
void merge(t){
	for_each(v in t.graph){
		edges = t.graph.get_edges(v);
		if(v not in this.graph) new_nbh[v] = move(edges);
		else
			new_nbh[v] = prune(
				edges + this.graph.get_edges(v),
				get_f_nbhs(this_label)); 	}
	ids.merge(t.ids); }
\end{lstlisting}

\fvamana{} is also based on \vamana{} with a variant of pruning algorithms. The algorithm relaxes the pruning conditions and additionally preserves a candidate $v$ as the final neighbor of the current vertex $u$ if $F_v \cup F_w \subseteq F_u$ where $w$ is a vertex already preserved. This logic is implemented by customizing the \textit{f\_nbhs($u,w$)} function as shown in \cref{code:fvamana}.
\begin{lstlisting}[
	language=C++,
	caption={Filtered Vamana Build},
	label={code:fvamana},
	morekeywords={foreach},
	escapechar=@
]
auto get_f_nbhs(seq labels){
	return [&](nid_t u){
		auto es = get_edges(u);
		auto check = filter([&](auto e){
			auto (u,v) = e;
			return u.label @$\cap$@ labels @$\neq \emptyset$@ });
		return es | check; 	}; }
\end{lstlisting}

The search for either \svamana{} or \fvamana{} is based on the regular \textit{beam search} by customizing the \textit{f\_nbhs} to filter for neighbors with specific labels.

\subsection{Filtered Search Evaluation}\label{app:filter}

\myparagraph{Dataset.} For filtered search, we utilize real-world datasets with native and synthetic labels.
\textit{MS MARCO Web Search (MARCO)}~\cite{chen2024ms} is an information-rich dataset incorporating ClueWeb22~\cite{overwijk2022clueweb22}'s high-quality web pages as its document corpus.
% The embedding vectors are accompanied by metadata such as semantic annotations and document ID.
We extract metadata to obtain 10 distinct labels based on the semantic annotations.
\textit{Yahoo Flickr Creative Commons (YFCC)}~\cite{thomee2016yfcc100m} comprises image embeddings from Flickr.
We evaluate the first 10M points with 200,363 distinct labels, following the setup of the NeurlIPS'23 Big-ANN Competition~\cite{simhadri2024results}.
Additionally, we randomly select 10M points from both \textit{BIGANN} and \textit{Cohere} and assign them synthetic labels in a Zipfian distribution ($\alpha=1$), which contains 50 distinct values.

To assess performance under varying restriction levels, we use a single label with different specificities, ranging from high to low percentiles, so that query difficulty spans from easy to more challenging.
During index construction, the filtering strategy identifies pairs of label sets with non-empty intersections, i.e., \(L_u \cap L_v \neq \emptyset\). 
At query time, the algorithm checks whether a point's label set contains the singleton query label, i.e., \(l \in L_u\).
For Milvus and Weaviate, we implement metadata filtering via the \textit{like} operation by concatenating each point's label set into a comma-separated string; these label strings serve as the metadata associated with each point. 
Milvus applies metadata filtering before ANN search by restricting the candidate set to entities that meet the filter, thereby improving performance. 
Weaviate supports hybrid search that combines keyword and vector search; users control the balance with \(\alpha \in [0,1]\). Because our experiments contain no keywords, we set \(\alpha=1\) to perform pure vector search with metadata filtering.

\begin{figure*}[!t]
  \centering
  \includegraphics[width=\linewidth]{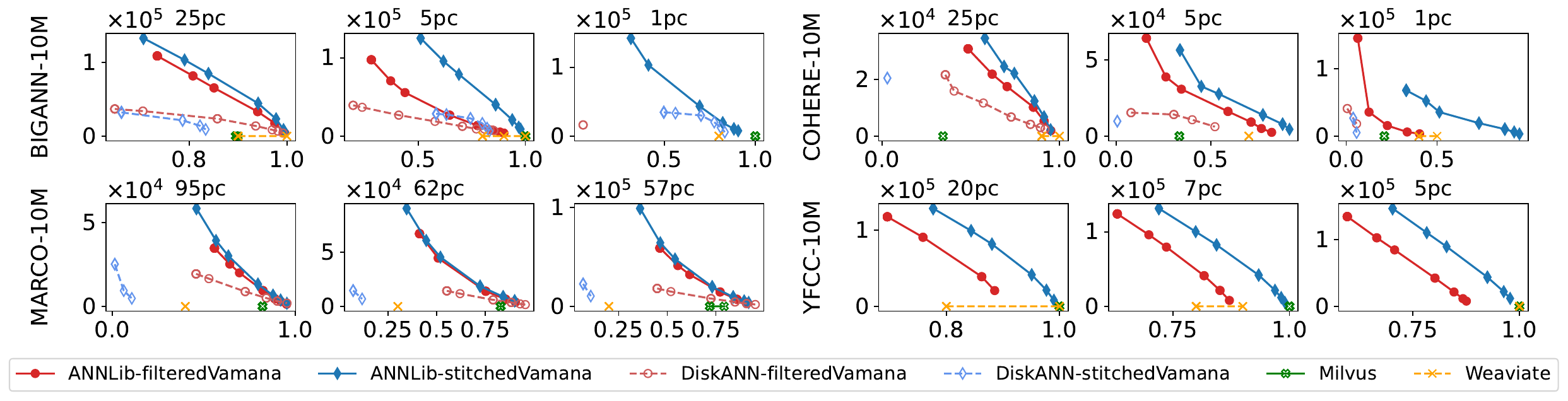}
  \caption{
  \textbf{QPS-recall of Filtered Search.}
  The throughput-recall curves performed on varying dataset and labels; x-axis for recall and y-axis for QPS.
  Each dataset uses three labels for query, indicating the specificity on the top.
  25pc means 25\% base points contains the queried label.
  \label{fig:filtered_search}
  \vspace{-0.5em}
}
\end{figure*}

%\subsection{Evaluating Filtered Search}\label{sec:eval:fsearch}
We present filtered search results for \annlib{} in comparison with the state-of-the-arts.
\Cref{fig:filtered_search} shows QPS--recall curves across four common vector datasets under varying specificity.
Here, specificity is defined as the fraction of points carrying a given label relative to the total number of points, where a lower value reflects more difficulty in the filtered search.
For each dataset, we select three specificities, shown as percent (pc), to evaluate the performance pattern.

\Cref{fig:filtered_search} shows \annlib{}'s competitive performance: it matches or outperforms the evaluated monolithic solutions in most cases, despite \annlib{}'s primary goal of simplifying ANNS system development.

\section{Batch Deletion}\label{app:bat-del-detail}
We implemented batch deletion with \annlib{} based on FreshDiskANN~\cite{singh2021fresh} where deletion simply marks the ones with a tombstone, and an extra \textit{consolidate} function is added to remove all the tombstones and recompute the incident edges from all two-hop-reachable neighbors. 
To avoid the significant computational costs of periodic index rebuilding, FreshDiskANN~\cite{singh2021fresh} proposed mark and consolidate methods to update index in place. Two steps of deletion that first marks points to be deleted into a delete list by running \textit{lazy\_delete}, and executes \textit{consolidate} method to remove the points in that delete list.
We set up a \texttt{state} array to record the current observed state. We maintain a global clock \texttt{tick} and monotonically increase its value after \texttt{erase}, indicating a new epoch of the deletion state.
At the point of traverse line \ref{line:del-mark-end}, the algorithm checks the existence of the vertex $v$ and marks it with the current \texttt{tick} for being alive or a special value \texttt{TOMB} as dead. The next time meeting with the current \texttt{tick}, the algorithm knows $v$ is alive and directly returns as a quick path.
% We additionally set a global time ticker to update the tombstone state lazily only when the node is visited. 
% Besides, the consolidation on a single point can also be triggered when 
\begin{lstlisting}[
	language=C++,
	caption={\vamana{} Batch Deletion},
	morekeywords={foreach},
	escapechar=@
]
auto get_f_nbhs(){
	return [&](nid_t u){
		auto es = get_edges(u);
		auto check = filter([&](auto e){
			auto (u,v) = e;
			if(state[v]==TOMB) return false;@\label{line:del-mark-begin}@
			if(state[v]==tick) return true;
			state[v] = id_map.contains(v)? tick: TOMB;@\label{line:del-mark-end}@
		});
		return es | check; 	};
}

void insert(auto pts){
	foreach(p : pts){ state[p] = tick; }
	... }

void erase(auto nids){
	id_map.delete(nids);
	tick++; }
\end{lstlisting}

\hide{
	linebackgroundcolor={
		\ifnum\value{lstnumber}=16
			\color{orange!30}
		\fi
		\ifnum\value{lstnumber}=22
			\color{orange!30}
		\fi
		\ifnum\value{lstnumber}=23
			\color{orange!30}
		\fi
		\ifnum\value{lstnumber}>5
			\ifnum\value{lstnumber}<9
				\color{orange!30}
			\fi
		\fi
	}

}

\subsection{Evaluating Deletion}
\label{sec:eval:del}

\begin{figure*}[t]
  \centering
  \includegraphics[width=\linewidth]{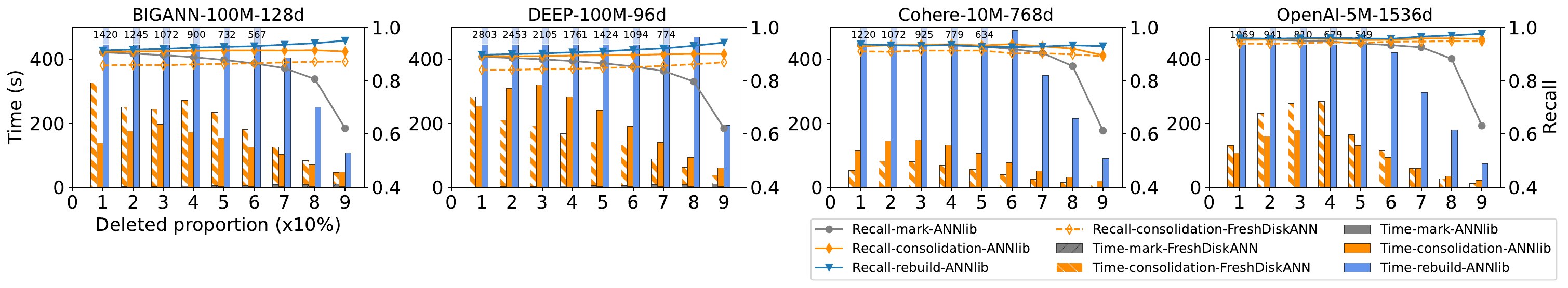}
  \caption{Deletion Time and Search Quality. Our main algorithm is ``consolidation-ANNLib''. ``FreshDiskANN'' is the original code from~\cite{singh2021fresh}. 
  Other solutions are baselines for comparison. ``Rebuild'' means to always fully rebuild the graph upon deletions. ``mark'' means to only mark the deleted points without removing them. 
  }
\end{figure*}

We also study the performance when batch deletion occurs.
Both \annlib{} and the baseline, \textit{DiskANN}, follow the method~\cite{singh2021fresh} of two stages, \textit{marking} and \textit{consolidation}.
% \textit{mark} all the deleted points, making them invisible during the search, and perform \textit{consolidation} at a proper point to physically remove the marked points and reconstruct the neighbors of the incident points.
In this test, we start with a fully built index and gradually remove points in batch, with 10\% of the original set.
We present the time taken for \textit{marking} and \textit{consolidation}.
Additionally, we provide the time to build an index from scratch on the remaining points for reference. To improve clarity, we clip the bars representing runtime over 250 seconds and annotate the actual numbers on the top. Alongside the time, we plot the query recall after each operation, under the same parameter settings, to indicate the index quality.

The time for \textit{marking} is linear to the index size and thus being trivial for small-size datasets. Simply \textit{marking} impairs the connectivity of the graph index, and the recall drops significantly as more points are marked.
\textit{Consolidation}, on the other hand, restores the connectivity and largely improves recall to the level of rebuilding, although being costly to evaluate $O(m^2)$ 2-hop neighbors. With more points being deleted, the time slightly rises at first and then decreases because the workload is positively related to $p\cdot n$ where $p$ is the proportion of the affected points small at the beginning, and $n$ is the number of remained points that is small at the end. Since \textit{consolidation} is costly, it dominates the trends of the whole deletion which consists of one \textit{marking} and one \textit{consolidation} in this test.

A full rebuild from the points is expensive. In our experiments, even when only 10\% points remained, the rebuild cost is still higher than a whole deletion, across varying dataset sizes and dimensions. However, one should note that, when sufficiently large numbers of points are deleted, it is possibly worth using rebuild as an alternative. The query recall slightly increases with fewer points remain because the fixed search parameters are more advantageous for a smaller scale.

\section{Range-Filtered Build Scalability}\label{app:range-scale}
\begin{figure*}[t]
\centering
\includegraphics[width=\linewidth]{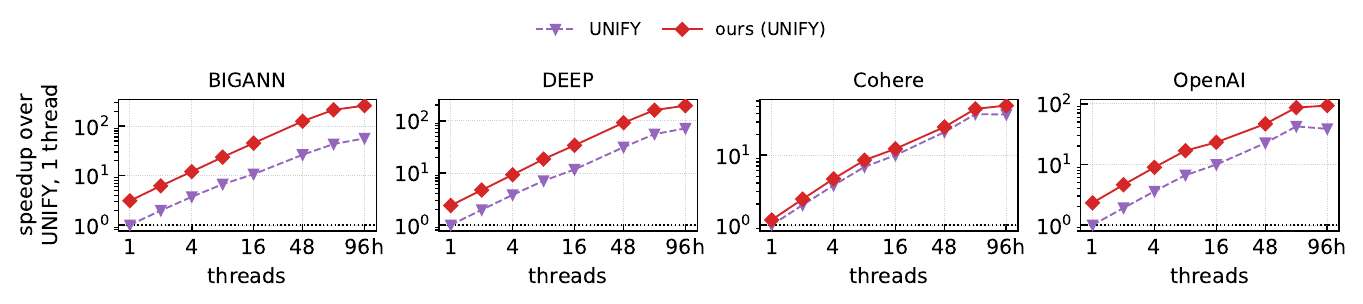}
\caption{\textbf{Build scalability of UNIFY against its \annlib{} re-implementation. Higher is better.} Both curves are speedup over the same reference, UNIFY on one thread, marked by the dotted line at $1$. 96h: 96 cores with hyperthreading.\label{fig:unify-scalability}}
\end{figure*}

\Cref{fig:unify-scalability} separates the two sources of the build-time gain reported for UNIFY in the main paper, by plotting both sides against the same reference, UNIFY on a single thread.
On one thread, where neither side has parallelism to exploit, the re-implementation is already $1.2$--$3.1\times$ faster.
This part of the gap is algorithmic, since a batch build performs the same per-point work without the synchronization the original inserts around every neighbor list it updates.
Parallelism then multiplies it: at the full machine the re-implementation reaches $52$--$259\times$ the single-thread baseline against UNIFY's own $38$--$71\times$.

The last step is the most direct evidence that lock contention, rather than the amount of work per point, is what limits the original.
Going from 96 cores to the full machine, UNIFY gets no further benefit from hyperthreading on \textit{Cohere} and \textit{OpenAI} and in fact regresses slightly, whereas the re-implementation continues to improve on both.
Hyperthreads share a core and therefore sharpen contention on the per-vertex locks, which penalizes the lock-mediated build exactly where the batch-parallel one still has room.

\section{Fine-granularity Customization}
\label{app:fine-gran-custom}
\begin{lstlisting}[
	language=C++,
	caption={Example of defining a custom tag},
	emph={custom\_tag\_parlay}
]
// parlay.hpp
class custom_tag_parlay{};
template<>
class custom<custom_tag_parlay>{
	...
};

// main.cpp
#include "parlay.hpp"
	namespace ANN::external{
	auto def_custom_tag(){
		return custom_tag_parlay{};
	}
}
\end{lstlisting}

\subsection{Support to uniform parallelism interfaces}
\annlib{} is designed not to depend on any third-party libraries while allows users to customize infrastructure, such as the parallel framework and memory allocator, by a few lines of code. This design makes industrial users easily adapt \annlib{} to their own products without modifying any library code and replace frequently-used functions for better performance in a clear form.

\subsection{Uniform fundamental types}
A library uses fundamental data types, such as \texttt{std::vector} or \texttt{boost::container::vector}, in its own code. The user code, however, may use a different type, e.g., \texttt{parlay::sequence}, to store the data. The deviation between the choices of such fundamental data types causes problems that may hinder the development: 1) the library can therefore depend on other third-party libraries, such as \texttt{boost}, and brings dependency management work or even conflicts; 2) the performance may be impaired due to the conversion between different types. For example, a user may insert a input dataset stored in \texttt{parlay::sequence} into the library, where the latter internally uses \texttt{std::vector} for storage, and a serial copy happens for the data transfer between the two container types. Ideally, if the library uses the same type as the user's, it shares the capability of parallel copy for better performance; moreover, in the case that the user do not use the input anymore, the input can be passed with a move semantics (e.g., by rvalue reference in C++) thus paying no cost on the data copy.

The design point is how to customize the provided library code and make it use the same fundamental data types as the user does. The fine-granularity customization mechanism in \annlib{} provides a clear design to allow the users to specify the fundamental types used in the library. \annlib{} is carefully written to work with the user-custom types with compatible interfaces. Therefore, it brings no burden of extra dependency management to the user and prevent unnecessary overhead caused by type conversion.

% \noindent\textbf{\textit{The \emph{to} helper}}

\subsection{Easy to debug}
Parallel algorithms, while providing higher performance, increase the difficulty in locating bugs due to extra efforts looking into multiple execution flows over threads and possible checking data races. A common practice is to serialize a suspicious piece of program and do debugging on it. The fine-granularity customization provides the ability to control the enablement of parallelism on any \annlib{} function or class by simply switching its custom tag to a serial one. The design eases the workload of bug locating in the parallelism environment, comparing to manually switching off all the parallel primitives in the involved code.

\subsection{Dual Reference-Counting Pointers / adapt to CPAM}

% \noindent\textbf{\textit{Separation between add\_nodes and set\_edges}}
\textit{customizable types}

\Cref{code:traverse} gives examples of parallel traverse over a CPAM tree and a \texttt{std::vector}.
% For containers that not support index-based parallel traverse
\begin{lstlisting}[
	caption={Filtering example},
	label={code:traverse},
	language=C++,
	emph={for\_each}
]
auto print = [&](const auto &e){
	std::print("{}", e);
};

struct cpam_wrap{
	using tree_t = cpam::pam_map<...>;
	tree_t map;

	void for_each(auto &&f){
		auto g = [&](const auto &e){
			f(e.first); // key
		};
		tree_t::foreach_index(map, g);
	}
};

cpam_wrap tree;
ANN::util::for_each(tree, print);

std::vector<int> seq;
// vector::{begin,end} are available
ANN::util::for_each(seq, print);
\end{lstlisting}

\begin{lstlisting}[
	caption={Filtering example},
	label={code:fnbhs-filter},
	language=C++
]
auto f_nbhs = [&](nid_t u){
	return filter_view(
		graph.get_edges(u),
		[&](nid_t v){
			return have_common_label(u,v);}
	);
};
\end{lstlisting}

\noindent\textbf{\textit{Reusable generic graph}}
% \noindent\textbf{\textit{}}
\begin{lstlisting}[language=C++,caption={Example of adapting a \texttt{hash\_map} to \annlib{}}]
// customize adj_map to use hashmap as the
// underlaying data structure
template<typename Key, typename T>
struct map_adapter : hashmap<Key,T>{
	...
	const T& operator[](const Key &key) const{
		return base::find(key)->second;
	}
};

template<class Nid, class Ext, class Edge>
using adj_hashmap = 
	ANN::adj_map<Nid, Ext, Edge, map_adapter>;
\end{lstlisting}

\section{Snapshot}

\begin{figure*}[h]
  \centering
  \includegraphics[width=\linewidth]{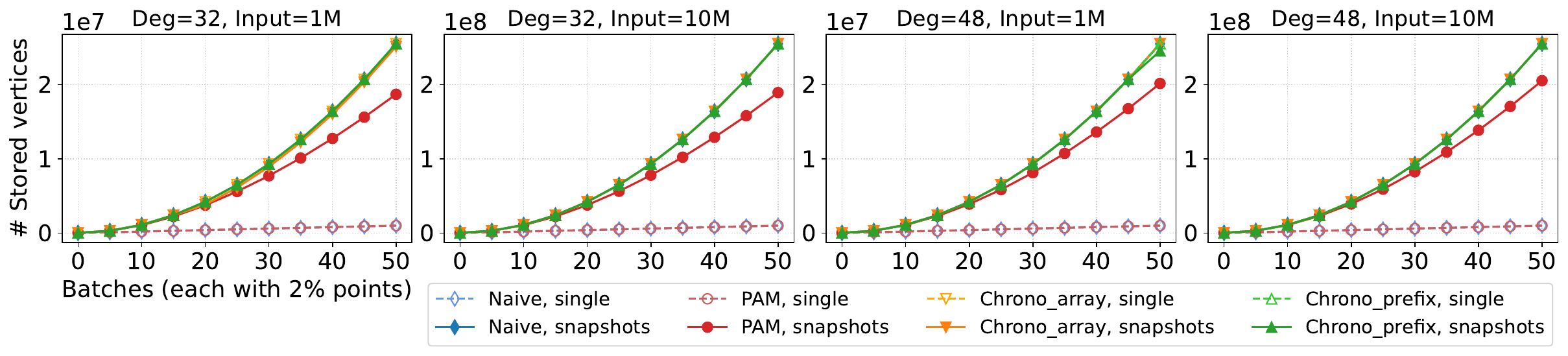}
  \caption{\#vertices stored in memory\label{fig:exp:app:snapshot1}}
\end{figure*}

\begin{figure*}[h]
  \centering
  \includegraphics[width=\linewidth]{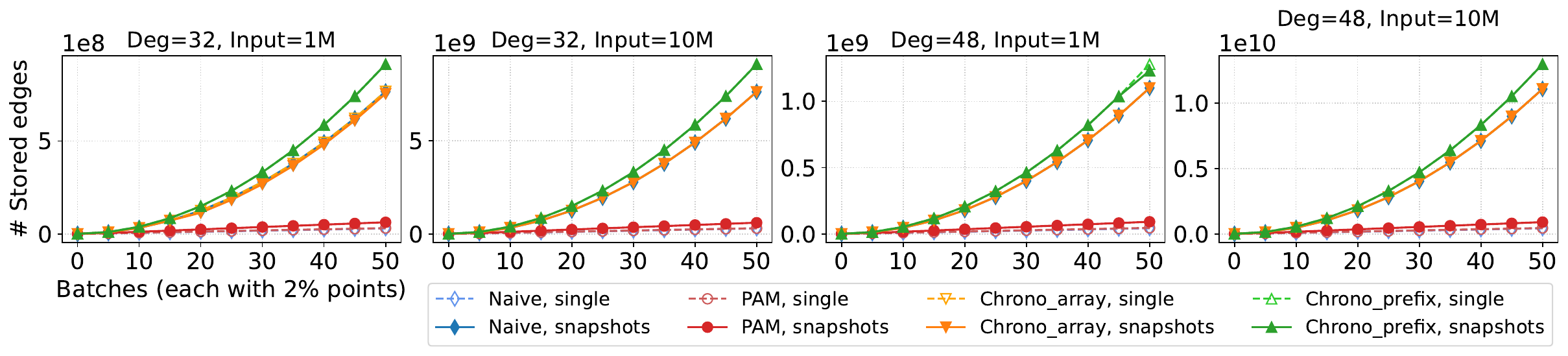}
  \caption{\#edges stored in memory\label{fig:exp:app:snapshot2}}
\end{figure*}

We evaluate the memory usage for different solutions for enabling snapshots, and report results in \cref{fig:exp:app:snapshot1,fig:exp:app:snapshot2}.

\section{Parallel Batch Insertion on \hnsw{}}\label{app:bat-ins-h}

\subsection{Algorithm and Implementation}
The parallel batch insertion of \hnsw{} on each layer is similar with \vamana{} except the upper layers only stores edges as connection information, and the bottom layer additionally stores the coordinates and height information and participates the distance computation. Such logic is achieved by modifying \fnbhs{} and \fdist{} as shown in \cref{code:ins-h}.
\begin{lstlisting}[
	language=C++,
	caption={\hnsw{} Batch Insertion},
	label={code:ins-h},
	morekeywords={foreach},
	escapechar=@,
	linebackgroundcolor={
		\ifnum\value{lstnumber}=2
			\color{orange!30}
		\fi
		\ifnum\value{lstnumber}=3
			\color{orange!30}
		\fi
		\ifnum\value{lstnumber}=7
			\color{orange!30}
		\fi
		\ifnum\value{lstnumber}=8
			\color{orange!30}
		\fi
	}
]
auto get_f_nbhs(level){
	return [&](nid_t u){@\label{line:app-ins-fnbhs-begin}@
		return g[level].get_edges(u);@\label{line:app-ins-fnbhs-end}@
	};
}
auto get_f_dist(nid_t v){
	return [&](nid_t u){@\label{line:app-ins-fdist-begin}@
		return dist(g[0].get_node(u), g[0].get_node(v));@\label{line:app-ins-fdist-end}@
	};
}
\end{lstlisting}

\section{Full Related Work}
\label{sec:related-work-full}

Here we present the full literature review, extending the discussion in \cref{sec:related-work}.
% , and only review the most relevant ones here. 

\myparagraph{ANNS Algorithms.}
There have been broad range of studies in ANNS algorithms such as DiskANN~\cite{subramanya2019diskann}, Stars~\cite{carey2022star}, SPFresh~\cite{xu2023spfresh}, ParlayANN~\cite{manohar2024parlayann}, and other graph-based approaches~\cite{malkov2020hnsw,fu2019nsg,munoz2019hcnng,fu2021high,mcinnes2020pynndescent,zhang2022hierarchical,ren2020hmann}. These state-of-the-art ANN systems can handle billion-scale datasets in a single node, while the codebase also grows with more than 10k lines of code~\cite{subramanya2019diskann}, making it complicated to find and modify related part to port features and tailor functions.

Many algorithms in \annlib{} are adapted from existing approaches. In particular, we support parallelism by prefix-doubling proposed by ParlayANN. 
%Among them, DiskANN consists of in-memory algorithm, \vamana{}, and the disk storage in a LSM-like manner. %It reduces the data transfer latency from the disk by both an adaptive search algorithm and the strategy only reading quantized vectors from the disk. 
ParlayANN~\cite{manohar2024parlayann} is a large-scale ANN benchmark framework that implements four prominent graph-based algorithms in a batch parallel manner. It enables billion-scale batch insertion in a single node by improving the scalability with the increase of thread. 
ParlayANN contains four graph-based algorithms, and encapsulates several general-purpose interfaces to interact with the graph structure. However, it assumes the existent of only one graph in the index, making it hard to adapt to multi-graph indices such as HNSW, HSSG, and HVS. Besides, the logic of beam search algorithm needs to be redesigned to adapt to filtering applications.

%Furthermore, to reduce the long build time of graph-based algorithms, the recent techniques resort to combine multiple techniques 
%in data pre-processing and space compressing. 
% We believe our library will be able to handle such hybrid cases and simplify the implementation of such sophisticated design by adding more abstracts and building blocks in the future.

% With the rapid advances of deep learning in the past decades, nowadays, the number of embedded vectors in training has grown to billions, and the dimensions of each vector can span to hundreds and thousands. To address the rising performance issue in large scales, ANNS is therefore proposed and studied.
%With multiple techniques, the state-of-the-art ANN systems can handle billion-scale datasets in a single node, while the codebase also grows with more than 10k lines of code~\cite{subramanya2019diskann}, making it complicated to find and modify related part to port features and tailor functions.

%Despite the needs to fast develop ANNS algorithms and flexibly extend the functionalities, 
%however, existing ANNS systems do not focus on such flexibility and the extensibility.

%ANNS frameworks are usually under active development thus varying code structure and function interfaces frequently. As a result, the developer need to carefully rewrite the code to implement the features in existing code while make extra efforts to cause any unexpected side effects to other functions. Due to the various techniques, however, it is naturally costly to both understand the old code and design the new features.

\myparagraph{Vector Databases.}
ANNS is also widely used in databases and serve vector queries. It is called vector databases for the capability to operate on vectors and perform ANNS. A database may be a traditional database equipped with vector query extension~\cite{pgvector,zhang2023vbase} or originally designed with ANNS support~\cite{wang2021milvus}. Either the extension or the original vector databases usually use single or a few ANNS algorithms as its core function, attached with the peripheral functions of the database. The process of wrapping the core ANNS algorithm into a database (extension) requires a deep modification to adapt to the database portion. 
% The database provides standard specifications of 
The common ANNS algorithms do not preserve the capabilities to communicate with external systems such as a database.

\myparagraph{Filter ANNS.} 
%There are a host of ANNS related applications, including filtered search, dynamic update, and out-of-distribution query. 
Filtered ANNS is an extension of ANN search by retrieving the top-k similar results while satisfying a specified set of metadata predicates.
% There are several existing work on filtered ANNS. We categorize recent filtered ANNS approaches in 2 categories by the ingredients of ANN indexes.
Existing works on filtered ANNS can be broadly classified into two main categories in this paper based on how they integrate the filtering mechanism with the main ANNS index. 
The first category is ad-hoc methods that treat filtering operation as an independent step, performed either before (pre-filtering) or after (post-filtering) the vector search, without modifying the original ANNS indexes.  
Pre-filtering first predicts on filtering conditions by building an independent index for each potential filters, then perform the vector search while evaluating the filters on candidates. Systems like ACORN~\cite{patel2024acorn}, Weaviate~\cite{weaviate2024}, Milvus~\cite{wang2021milvus}, AnalyticDB-V~\cite{wei2020analyticdb}, SingleStore-V~\cite{chen2024singlestore}, and MSVBASE~\cite{zhang2023vbase} adopt this approach. 
However, because it is hard to predict the filtering patterns and elements that will be contained in the search, this approach scales poorly, becoming inefficient for higher filtering labels on large datasets. 
By contrast, post-filtering first runs a standard ANN query to retrieve a larger candidate set than the target $k$. Then, it scans this set to identify those that match the specified attributes. This technique, implemented in systems like PASE~\cite{yang2020pase} and FAISS~\cite{douze2024faiss, johnson2019billion}, is straightforward to deploy. However, it becomes highly inefficient when the filter is very selective because the vast majority of retrieved candidates are discarded, resulting in significant wasted computation.

The second category involves building filter-aware indexes that integrate filtering information into the ANN indexes. This approach modifies the index construction and search algorithms to simultaneously handle vector similarity and metadata constraints, as in Filtered-DiskANN~\cite{gollapudi2023filtered}, or to combine the attribute constraints with the indexes, such as Native hybrid query (NHQ)~\cite{wang2023efficient} and MA-NSW~\cite{xu2020multiattribute}. IVF$^2$ introduces a label-centric IVF-based (inverted file indexing) approach to apply the Vamana index and an IVF index according to the query specificity.

Recent research has expanded the application of filtered ANNS to related problems and platforms. For instance, several studies have adapted these techniques to accelerate filtered range queries~\cite{engels2024approximate, zuo2024serf}. Meanwhile, \cite{lin2025survey} provides a framework to comprehensively summarize the latest developments of filtered ANNS. Furthermore, novel indexing and search optimizations have been proposed to enhance the performance of filtered ANNS on specialized hardware like GPUs~\cite{xi2025vecflow}.

\myparagraph{Dynamic ANNS.} To enhance the scenario of dynamically updating and deleting data in real-world ANNS systems, Fresh-DiskANN~\cite{singh2021fresh} has been introduced to achieve the deletion by employing two-pass lazy-delete and consolidate instead of periodically reconstructing existing indexes. At the same time, it maintains the navigability of the indexes and the search performance on updated graphs.

%On the other hand, OOD-DiskANN~\cite{jaiswal2022ood} is specifically designed to efficiently handle out-of-distribution (OOD) queries. It incorporates a sparse set of OOD query samples during index construction and improves query-aware production quantization (PQ) to adapt to the target query distribution.

% \subsubsection{Heterogeneous}
% To improve the performance of ANNS, people have found that using heterogeneous devices accelerates the query process. GPUs are a popular way to speed up ANNS in parallel...
% Disks, such as SSDs, are also commonly used to store and search huge graphs...
% Hierarchical storage structures have also been chosen to solve various query scenarios...

\myparagraph{Benchmarks}
There are various benchmarks for evaluating the performance of ANNS using vector datasets. In this paper, we selected some of the following datasets as metrics. 

The primary challenge is scalability, addressed by billion-scale benchmarks. Among these, BIGANN-1B \cite{annbench, amuller2020ann, amsaleg2010datasets} and Deep-1B \cite{babenko2016efficient} are two benchmarks that widely used to measure the query quality and ANNS algorithm scalability. 
This focus on scale extends to specific applications, where Microsoft SpaceV-1B \cite{spacev2021spacev} provides document and query vectors to capture generic intent representation for both, while Microsoft Turing ANNS uniquely assesses query-to-query similarity.

Other datasets focus on simulating complex, real-world search applications. MS MARCO Web Search \cite{chen2024ms} reproduces the highly skewed distributions of real web documents and user queries, providing a realistic environment to evaluate performance on filtered search or out-of-distribution (OOD) queries. Similarly, the YFCC-100M \cite{thomee2016yfcc100m} pairs CLIP image embeddings \cite{radford2021learning} with rich metadata labels derived from text, creating a key test for filtered search where queries must satisfy both semantic and attribute constraints. The cross-modal Text-to-Image-1B (T2I) dataset \cite{baranchuk2021benchmarks} specifically challenges algorithms with OOD queries \cite{jaiswal2022ood}, where the query distribution (text embeddings) fundamentally differs from the indexed data (image embeddings). Finally, the Cohere Wikipedia dataset \cite{cohere_wikipedia_embeddings} provides high-dimensional vectors from a state-of-the-art multilingual model, serving as a critical benchmark for complex semantic search scenarios.

\end{document}